\pdfoutput=1

\documentclass[11pt]{article}

\usepackage[preprint]{acl}

\usepackage{times}
\usepackage{latexsym}

\usepackage[T1]{fontenc}

\usepackage[utf8]{inputenc}

\usepackage{microtype}

\usepackage{inconsolata}

\usepackage{booktabs}
\usepackage{comment}
\usepackage{enumitem}
\usepackage{graphicx}
\usepackage[table,dvipsnames]{xcolor}
\usepackage{multirow}
\usepackage{amsfonts}
\usepackage{amsmath}
\usepackage{xspace}
\usepackage[noabbrev, capitalize]{cleveref}
\usepackage{tikz}
\usepackage{pgfplots}
\pgfplotsset{compat=1.18}

\usepackage{float}

\usepackage{pdflscape}

\usepackage{rotating}

\definecolor{crimson}{RGB}{171, 47, 65}
\definecolor{forestgreen}{RGB}{60, 120, 60}
\definecolor{goldenrod}{RGB}{218, 165, 32}

\newcommand{\yago}{YAGO21K-610\xspace}
\newcommand{\grail}{GraIL\xspace}
\newcommand{\tyler}{TyleR\xspace}
\newcommand{\llama}{Llama\xspace}

\newcommand{\llamathreeb}{Llama3-8B\xspace}
\newcommand{\roberta}{RoBERTa\xspace}
\newcommand{\robertal}{RoBERTa-L\xspace}
\title{Type-Less yet Type-Aware Inductive Link Prediction with Pretrained Language Models\thanks{This is the authors’ version of the work. The final,
published version will appear in the \textit{Proceedings of the 2025 Conference on Empirical Methods in Natural Language Processing (EMNLP ‘25)}.
\\\\
This work is licensed under a Creative Commons Attribution 4.0 International License (CC BY 4.0).
\\\\
Please cite the official published version when available.}}

\author{
  \textbf{Alessandro De Bellis\textsuperscript{1}},
  \textbf{Salvatore Bufi\textsuperscript{1}},
  \textbf{Giovanni Servedio\textsuperscript{1,2}},
  \textbf{Vito Walter Anelli\textsuperscript{1}},
  \\
  \textbf{Tommaso Di Noia\textsuperscript{1}},
  \textbf{Eugenio Di Sciascio\textsuperscript{1}}
\\
\\
  \textsuperscript{1}Politecnico di Bari, Italy,
  \textsuperscript{2}Sapienza University of Rome, Italy
\\
  \small{
    \textbf{Correspondence:} \href{mailto:alessandro.debellis@poliba.it}{name.surname@poliba.it}
  }
}

\begin{document}
\maketitle
\begin{abstract}
Inductive link prediction is emerging as a key paradigm for real-world knowledge graphs (KGs), where new entities frequently appear and models must generalize to them without retraining. Predicting links in a KG faces the challenge of guessing previously unseen entities by leveraging generalizable node features such as subgraph structure, type annotations, and ontological constraints. However, explicit type information is often lacking or incomplete. 
Even when available, type information in most KGs is often coarse-grained, sparse, and prone to errors due to human annotation.
In this work, we explore the potential of pre-trained language models (PLMs) to enrich node representations with \textit{implicit} type signals. We introduce \textit{TyleR}, a \underline{Ty}pe-\underline{le}ss yet type-awa\underline{R}e approach for subgraph-based inductive link prediction that leverages PLMs for semantic enrichment. Experiments on standard benchmarks demonstrate that TyleR outperforms state-of-the-art baselines in scenarios with scarce type annotations and sparse graph connectivity. To ensure reproducibility, we share our code at \url{https://github.com/sisinflab/tyler}.
\end{abstract}

\section{Introduction}
\label{sec:introduction}
Knowledge graphs (KGs) represent complex relationships between entities in a structured, graph-based format~\cite{10.1145/3447772}. Their ability to encode semantic information and support reasoning makes them valuable in a variety of applications, such as natural language processing~\cite{DBLP:conf/emnlp/PetersNLSJSS19}, recommendation systems~\cite{DBLP:conf/www/WangSWX24}, and biomedical research~\cite{DBLP:journals/corr/abs-2305-19979}. 
However, KGs are notoriously incomplete: many valid relations are absent, reducing their effectiveness in downstream tasks \cite{DBLP:journals/tkdd/RossiBFMM21}. 

Link prediction aims to infer these missing relationships by analyzing the existing graph's structure and patterns.
Traditional link prediction methods aim to predict links among entities observed during training. Although effective in static settings, they are limited in dynamic environments where new entities are incrementally introduced.
Inductive link prediction (ILP) addresses this challenge by aiming to generalize to previously unseen entities, leveraging transferable features such as structural information and type information.

Prior work has demonstrated that incorporating entity type information can enhance generalization capabilities of ILP models.
For instance, \citet{zhou-etal-2023-inductive} explicitly integrate type annotations and ontological constraints into the learning process. 
Yet, these methods face a critical bottleneck: available explicit type information in real-world KGs is often coarse-grained, incomplete, or even erroneous. This limitation is particularly acute when facing structural sparsity. 
Consider, for instance, the triple \texttt{\textlangle Lionel Messi, playedFor, Barcelona FC\textrangle}. A model might assign similar plausibility to \texttt{\textlangle Cristiano Ronaldo, playedFor, Barcelona FC\textrangle} if both subject entities (i.e., \texttt{Lionel Messi} and \texttt{Cristiano Ronaldo}) lack distinct neighborhood information and are categorized only under the broad type "\texttt{Footballer}".
This highlights a fundamental inadequacy of type-informed ILP approaches when explicit type signals are weak and local graph structure is uninformative.

To address this gap, our idea is to leverage the rich semantic knowledge captured by pre-trained language models (PLMs). 
We hypothesize that the semantic understanding these models acquire during their extensive pre-training on vast textual corpora \cite{petroni-etal-2019-language, DBLP:conf/acl/HaoTTNSZXH23} offers a pathway to a more fine-grained representation of entities.
This "inner knowledge," encoded within the PLM's parameters, offers a dense representation of diverse semantic facets.
For example, prompting a PLM like BERT \cite{DBLP:conf/naacl/DevlinCLT19} with "Paris is located in \underline{\hspace{10pt}}," generates a hidden representation for the missing token that (ideally) enables it to correctly predict "France," reflecting the model's "\textit{understanding}" of Paris's geographical location.
We aim to utilize the implicit semantic insights of PLMs to derive fine-grained entity representations, overcoming limitations in explicit type information.
We start from these two observations: (i) an entity can be described by a set of assertions defining its properties; (ii) the same assertions, when used as prompts for a PLM, can elicit dense, multifaceted representations that implicitly capture a "type-aware" understanding of the entity. 
This potential led us to ask: \textit{Can PLM-derived entity representations compensate for structural and type sparsity in inductive knowledge graph completion?}
To investigate this question, we introduce \textbf{TyleR}--\emph{\textbf{Ty}pe-\textbf{le}ss yet type-awa\textbf{R}e}--a novel inductive link prediction framework that leverages PLMs to embed implicit type-aware signals within node representations, thus eliminating reliance on explicit type annotations.
Our contributions are:
\begin{enumerate}
\item We introduce a novel methodology for harnessing PLMs to derive and embed implicit type semantics within an ILP model, thereby enabling nuanced entity representations without relying on explicit type data.
\item We demonstrate TyleR's effectiveness on multiple benchmark datasets, showing its capability to perform competitively, especially in settings with limited or coarse-grained type information and sparse graph structures.
\item 
We conduct an empirical analysis investigating the interplay between PLM-derived semantic features and varying levels of type and structural sparsity, thereby characterizing the resilience of our approach.
\end{enumerate}



The remainder of the paper is organized as follows: \cref{sec:background} introduces the idea behind subgraph-based relational inference; \cref{sec:approach} details the methodology; \cref{sec:experimental-setup} describes the experimental setup and evaluation; \cref{sec:results} presents the results; \cref{sec:related} reviews related work; and \cref{sec:conclusion} concludes with future directions.


\section{Background and Motivation}
\label{sec:background}

Inductive link prediction aims to predict the likelihood of triples $(h, r, t)$, where $h$ and $t$ are unseen entities. In practice, this is done by means of a scoring function $f(h, r, t)$. At training time, $f$ is optimized on the triples in a training graph $\mathcal{G}_{train}$. At test time, the same scoring function is used to predict the plausibility of triples $(h', r, t')$ belonging to a test graph $\mathcal{G}_{test}$, based on the triples in an inference graph $\mathcal{G}_{inf}$.
Unlike traditional embedding-based approaches, subgraph-based relation prediction methods such as \grail~\cite{pmlr-v119-teru20a} can be viewed as learning logical rules that capture entity-independent relational semantics. For example, one can derive the simple rule:
\begin{equation*}
\medmuskip=0mu
\thinmuskip=0mu
\thickmuskip=0mu
\begin{split}
    \textit{spouse\_of}(X, Y) \land \textit{lives\_in}(Y, Z) \rightarrow \textit{lives\_in}(X, Z). 
\end{split}
\label{eq:rule1}
\end{equation*}

As demonstrated by~\citet{zhou-etal-2023-inductive}, the reasoning capabilities of \grail can be enhanced by incorporating explicit \textit{type information} about entities. This additional semantic context enables the model to induce more precise and type-aware rules:
\begin{equation*}
\medmuskip=-2mu
\thinmuskip=-2mu
\thickmuskip=-2mu
\begin{split}
\textit{Employee}(X) &\land \textit{Department}(Y) \land \textit{Office}(Z) \land \\
\land \textit{part\_of}(X, Y) &\land \textit{located\_in}(Y, Z)
\rightarrow \textit{works\_in}(X, Z).
\end{split}
\end{equation*}

Type-constrained rules enhance both accuracy and interpretability in relational inference by reducing spurious predictions and enforcing semantic validity. However, explicit type information is often incomplete or missing in real-world knowledge graphs. To address this, we propose learning a function \(\tau_{PLM}\), parameterized by a pre-trained language model, that maps entities to implicit type representations capturing their latent semantics. These PLM-derived embeddings enable type-aware reasoning without explicit type labels and can be integrated into the logical rule induction process. For example, a type-aware rule may take the form:
\begin{equation*}
\medmuskip=0mu
\thinmuskip=0mu
\thickmuskip=0mu
\begin{split}
\tau_{\text{PLM}}(X) 
&\land \tau_{\text{PLM}}(Y) 
\land \tau_{\text{PLM}}(Z) \land 
 \textit{part\_of}(X, Y) \land\ \\ &\land\ \textit{located\_in}(Y, Z)
\rightarrow \textit{works\_in}(X, Z),
\end{split}
\end{equation*}
with \(\tau_{\text{PLM}}(X)\), \(\tau_{\text{PLM}}(Y)\), and \(\tau_{\text{PLM}}(Z)\) such that
\begin{equation*}
\begin{split}
\tau_{\text{PLM}}(X) &\approx \textit{Employee}(X), \\ 
\tau_{\text{PLM}}(Y) &\approx \textit{Department}(Y), \\
\tau_{\text{PLM}}(Z) &\approx \textit{Office}(Z),
\end{split}
\end{equation*}
where $\tau_{\text{PLM}}(\cdot)$ for $X$ is an approximation of the logical statement $\textit{Employee}(\cdot)$ while, for $Y$ and $Z$, $\tau_{\text{PLM}}(\cdot)$ is an approximation of their types, \textit{Department} and \textit{Office}, respectively
(more details in~\cref{sec:approach}). This guides the rule induction process towards more meaningful and generalizable patterns, allowing us to infuse latent type semantics into subgraph-based link prediction models, even when explicit type information is absent.

\section{Methodology}
\label{sec:approach}

\begin{figure*}[t]
    \centering
  \includegraphics[width=\linewidth]{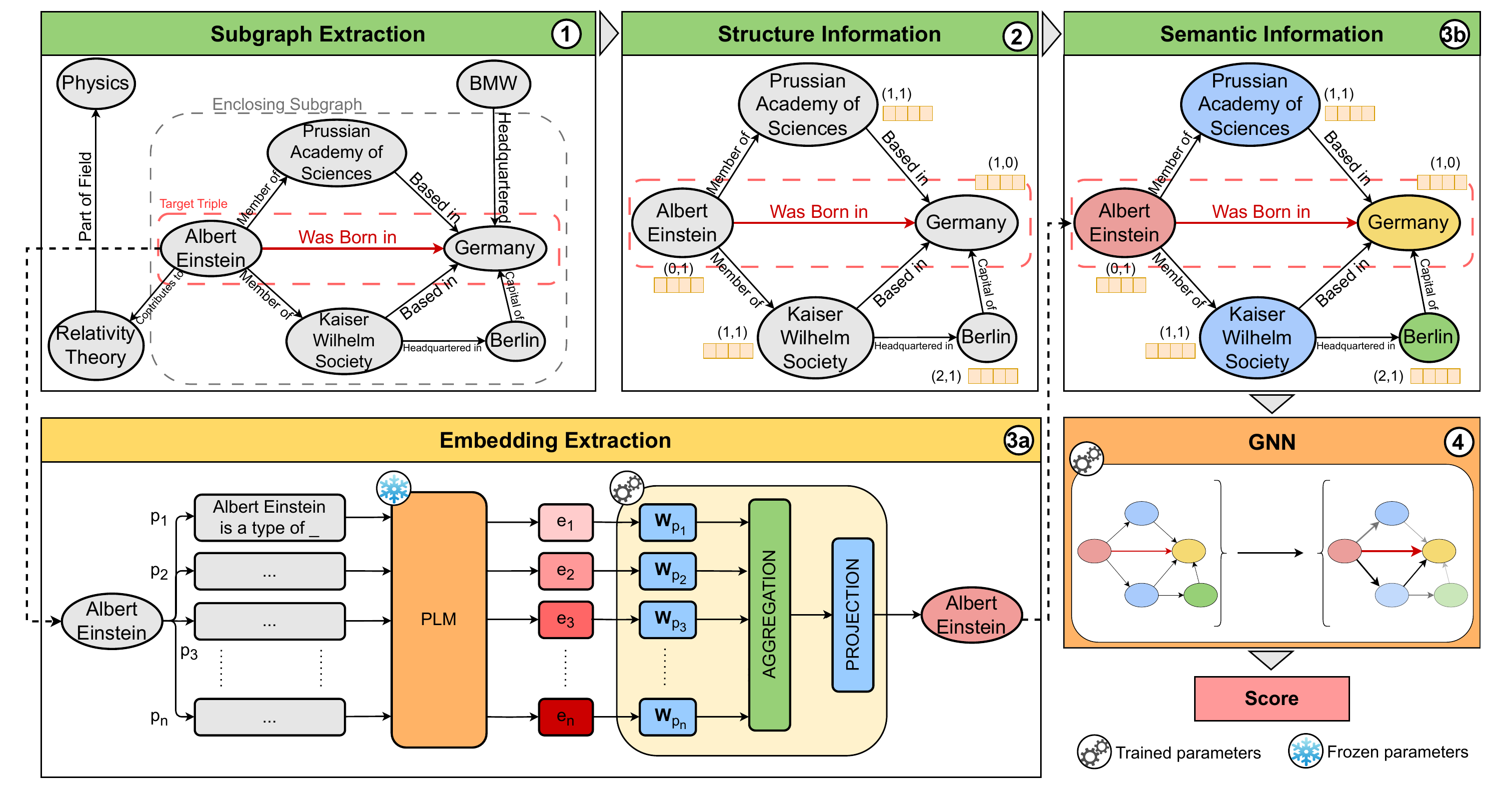} \hfill
    \caption{Overview of \tyler. 
    The process begins with \textcircled{\raisebox{-0.5pt}{\footnotesize\textbf{1}}} extracting the enclosing subgraph and \textcircled{\raisebox{-0.5pt}{\footnotesize\textbf{2}}} applying a node labeling strategy. Multi-faceted, semantic representations are then derived using a pre-trained language model \textcircled{\raisebox{-0pt}{\scriptsize\textbf{3a}}}, \textcircled{\raisebox{-0pt}{\scriptsize\textbf{3b}}}. Finally, a graph neural network \textcircled{\raisebox{-0.5pt}{\footnotesize\textbf{4}}}
    integrates structural and semantic information to obtain the final prediction.}
    \label{fig:tyler-diagram}
\end{figure*}

In this section, we introduce \textbf{\tyler} (\textbf{Ty}pe-\textbf{le}ss yet type-awa\textbf{R}e inductive link prediction with pre-trained language models). Building on Graph Inductive Learning~\cite{pmlr-v119-teru20a}, which infers relations from local subgraph patterns, \tyler leverages PLM-derived semantics to enrich node representations. However, integrating PLMs into full-graph models is computationally expensive due to high-dimensional embeddings and large graph size. \tyler adopts a subgraph-reasoning approach, restricting triple scoring to compact and informative subgraphs, making PLM integration tractable.

As illustrated in \cref{fig:tyler-diagram}, \tyler’s pipeline consists of four stages: \textcircled{\raisebox{-0.5pt}{\footnotesize\textbf{1}}} extracting an enclosing subgraph, \textcircled{\raisebox{-0.5pt}{\footnotesize\textbf{2}}} structurally labeling nodes (following GraIL~\cite{pmlr-v119-teru20a}), \textcircled{\raisebox{-0pt}{\scriptsize\textbf{3a}}}, \textcircled{\raisebox{-0pt}{\scriptsize\textbf{3b}}} enriching nodes with PLM-based semantic embeddings, and \textcircled{\raisebox{-0.5pt}{\footnotesize\textbf{4}}} feeding the enhanced subgraph into a GNN architecture from \citet{zhou-etal-2023-inductive}. The following sections provide further details on each step.



\paragraph{Subgraph Extraction \textcircled{\raisebox{-0.5pt}{\footnotesize\textbf{1}}}} 

Given a target triple $(u, r_t, v)$, we define $\mathcal{N}_k(u)$ and $\mathcal{N}_k(v)$
as the sets of $k$-hop neighboring nodes of $u$ and $v$, respectively. We also define a specific distance metric \( d(i, u) \) as the shortest path from a node \( i \) to \( u \) that does not pass through \( v \), and \( d(i, v) \) is similarly the shortest path distance from \( i \) to \( v \) that does not pass through \( u \).
The \textit{enclosing subgraph} of triple $(u, r_t, v)$ is computed by (i) forming an initial set of candidate nodes by taking the intersection $\mathcal{N}_k(u) \cap \mathcal{N}_k(v)$ and (ii) pruning nodes that are either isolated (i.e., have no edges connecting it to other nodes within the subgraph after this pruning step) or for which $d(i,u) > k$ or $d(i,v) > k$. The remaining nodes and their edges form the enclosing subgraph.

\paragraph{Subgraph Labeling \textcircled{\raisebox{-0.5pt}{\footnotesize\textbf{2}}}}

Each node \( i \) in the extracted subgraph is labeled with a pair of shortest path distances \( (d(i, u), d(i, v)) \) to the target nodes $u$ and $v$, respectively, within the subgraph. This pair captures the relative \textit{position} of node \( i \) with respect to the target nodes \( u \) and \( v \). The final positional embedding \( \mathbf{h}^{\text{pos}}_i \) is:
\begin{equation}
    \mathbf{h}^{\text{pos}}_i = \text{one-hot}(d(i, u)) \oplus \text{one-hot}(d(i, v)),
\end{equation}
where $\oplus$ denotes the concatenation operator and one-hot$(\cdot)$ is the one-hot encoding function. All nodes in the enclosing subgraph are within \( k \) hops of \( u \) or \( v \), so $ \mathbf{h}^{\text{pos}}_i \in \mathbb{R}^{2k + 2} $.

\paragraph{Semantic Enrichment \textcircled{\raisebox{-0pt}{\scriptsize\textbf{3a}}}\textcircled{\raisebox{-0pt}{\scriptsize\textbf{3b}}}}
\label{sec:semantic-enrichment}
Semantic enrichment leverages a pre-trained language model (PLM), supporting either masked token prediction or next-token generation.
A straightforward approach for encoding entity type semantics involves prompting the PLM with an explicit query to elicit the most plausible type for a given entity.
For masked language models (MLMs) (e.g., \roberta), this operation results in a prompt such as \textit{``The type of Paris is \texttt{[MASK]}''}, with the type semantics encoded in the last hidden layer representation of the \texttt{[MASK]} token; for causal language models (CLMs) (e.g., \llama), this representation corresponds to the last hidden representation of the final sequence token.
However, relying solely on representations derived from such direct type queries can be suboptimal.
Prior research has shown that transformer-based representations tend to be highly anisotropic, often concentrated in narrow cones~\cite{DBLP:conf/emnlp/Ethayarajh19}, which can limit their discriminative utility. To address this, we propose to refine type semantics through multiple prompts, designed to extract different semantic aspects. Given an entity \( i \) with textual label \( l_i \), we define a set of \textbf{assertion prompts} \( P = \{ p_1, p_2, \dots, p_n \} \), where each \( p_k \) targets a semantic facet of the entity (e.g., type, location, membership). Each prompt \( p_k(l_i) \) is processed by the PLM (\textcircled{\raisebox{-0pt}{\scriptsize\textbf{3a}}}) to yield a latent representation:
\begin{equation}
    \mathbf{z}_{p_k,i} = \text{Extract}(\text{PLM}(p_k(l_i))).
\end{equation}
\noindent
Here, $\text{PLM}(\cdot)$ denotes the forward pass of the language model given an input prompt, and $\text{Extract}(\cdot)$ selects the relevant hidden state (i.e., the \texttt{[MASK]} token’s final hidden layer for MLMs or the last token’s representation for CLMs). 
These representations \( \mathbf{z}_{p_k, i} \) are refined and projected into a unified space using an assertion-specific projection block:
\begin{equation}
    \mathbf{z}_{p_k, i}^h = \mathbf{W}_{p_k} \text{LN}(\mathbf{z}_{p_k,i}) + \mathbf{b}_{p_k},
\end{equation}
\noindent
where $\mathbf{W}_{p_k}$ and $\mathbf{b}_{p_k}$ are specific learnable parameters for each assertion prompt $p_k$, and $\text{LN}$ denotes layer normalization~\cite{DBLP:journals/corr/BaKH16}.
We aggregate ($\text{AGG}_{p}(\cdot)$) the prompt representations with different strategies such as \textit{sum}, \textit{mean} or \textit{concatenation}:
\begin{equation}
\label{eq:prompt-agg}
\mathbf{z}_i^\text{agg} = \text{AGG}_{p}( \{ \mathbf{z}_{p_k, i}^h\}_{k=1}^{n} ).
\end{equation}
\noindent
The semantic embedding $h^{sem}_i$ is obtained as:
\begin{equation}
    \mathbf{h}^{sem}_i \equiv \tau_{\text{PLM}}(i) =\sigma(\mathbf{W}_o\text{ReLU}(\mathbf{z}^\text{agg}_i)) ,
\end{equation}
where $\tau_{\text{PLM}}(i)$ is a function capturing the semantics of $i$ by aggregating multiple prompt-based representations via a PLM (as introduced in~\cref{sec:background}), and $\sigma(\cdot)$ is the sigmoid function.
Given a node $i$, we then construct the embedding $h_i^0$ as (\textcircled{\raisebox{-0pt}{\scriptsize\textbf{3b}}}):
\begin{equation}
    \mathbf{h}^0_i = [\mathbf{h}^{pos}_i \oplus \mathbf{h}^{sem}_i ].
\end{equation}

\paragraph{GNN Scoring \textcircled{\raisebox{-0.5pt}{\footnotesize\textbf{4}}}}

As suggested by~\citet{zhou-etal-2023-inductive}, our base GNN follows the R-GCN~\cite{DBLP:conf/esws/SchlichtkrullKB18} architecture. At layer $l$, the embedding for a node $i$ is computed as:
\begin{equation}
    \mathbf{h}_i^{(l)} = \text{ReLU}(\mathbf{W}_0^{(l)}\mathbf{h}^{(l-1)}_i + \mathbf{a}^{(l)}_i),
\end{equation}
\noindent
where $\mathbf{W}_0^{(l)}$ is a self-loop learnable matrix and $\mathbf{a}_i^{(l)}$ is the AGGREGATE function, based on edge attention~\cite{pmlr-v119-teru20a} and entity-relation composition~\cite{DBLP:conf/iclr/VashishthSNT20}:
\begin{equation}
\begin{split}
     \mathbf{a}^{(l)}_i =\sum_{r\in R} \sum_{j \in \mathcal{N}^r(i)} \alpha^{(l)}_{rr_tji} \mathbf{W}_r^{(l)} (\mathbf{h}_j^{(l-1)} - \mathbf{e}_r^{(l-1)}),
\end{split}
\end{equation}
\noindent
where $\mathbf{W}_r^{(l)}$ is a relation-specific transformation matrix at layer $l$, $\mathcal{N}^r(i)$ is the set of outgoing neighboring nodes of node $i$ under relation $r$. We adopt basis sharing~\cite{DBLP:conf/esws/SchlichtkrullKB18} as regularization for the $\mathbf{W}_r^{(l)}$ transformation matrices, whereas $\mathbf{e}_r^{(l)}$ is the relation embedding at layer $l$:

\begin{equation}
    \mathbf{e}^{(l)}_r = \mathbf{W}^{(l)}_{rel} \mathbf{e}^{(l-1)}_r.
\end{equation}
The edge attention weight $\alpha^{(l)}_{rr_tji}$ quantifies the importance of an edge $(j, r, i)$ when inferring relation $r_t$ at layer $l$.
\begin{equation}
    \alpha^{(l)}_{rr_tji} = \sigma(\mathbf{W}_\alpha^{(l)}\mathbf{s}^{(l)}_{rr_tji} + b_\alpha^{(l)}),
\end{equation}
\begin{equation}
\begin{split}
   \mathbf{s}^{(l)}_{rr_tji} = \text{ReLU}(\mathbf{W}_s^{(l)}[\mathbf{h}_j^{(l-1)}\oplus \mathbf{h}_i^{(l-1)}\\\oplus\ \mathbf{e}_r^{(l-1)}\oplus \mathbf{e}_{r_t}^{(l-1)}] + \mathbf{b}_s^{(l)}).
\end{split}
\end{equation}
To obtain the final representation of a node, \citet{pmlr-v119-teru20a} suggest adopting JK-Connections~\cite{DBLP:conf/icml/XuLTSKJ18}, i.e., by concatenating all the intermediate-layer representations. After the aggregation, the final score is computed as
\begin{equation}
\medmuskip=-1mu
\thinmuskip=-1mu
\thickmuskip=-1mu
\begin{split}
f(u, r_t, v) = \mathbf{W}_f^T \bigoplus_{l=1}^L[\mathbf{h}^{(l)}_\mathcal{G}(u, r_t, v) \oplus\ \mathbf{h}^{(l)}_u \oplus \ \mathbf{h}^{(l)}_v  \oplus \ \mathbf{e}^L_{r_t}],
\end{split}
\end{equation}
\noindent
where $\mathbf{h}^{(l)}_\mathcal{G}(u, r_t, v)$ is the subgraph representation, obtained via average pooling over all node representations at level $l$ in the subgraph.

\paragraph{Loss Function}

We adopt a margin-based pairwise loss function, which aims at maximizing the score on positive triples and minimizing the score on randomly sampled negative triples:
\begin{equation}
\medmuskip=0mu
\thinmuskip=0mu
\thickmuskip=0mu
\begin{split}
\mathcal{L} = \sum_{(u, r_t, v) \in G}max(0, f_e(u', r_t, v') -  f_e(u, r_t, v) + \gamma),
\end{split}
\end{equation}

\noindent where $\gamma$ is a margin hyperparameter, $(u, r_t, v)$ is a positive triple and $(u', r_t, v')$ is a negative triple.

\section{Experimental Setup}
\label{sec:experimental-setup}
In this section we detail our experimental setup, including datasets, baselines, training and evaluation details. Experiments were conducted with Python 3.8.19 and PyTorch 2.3.0, using an NVIDIA Ampere A100 GPU (64GB VRAM) and CUDA 12.1.

\subsection{Datasets}

We conduct experiments on \yago~\cite{zhou-etal-2023-inductive} and three FB15K-237 (FB237 in short) variants (v1–v3) from \citet{pmlr-v119-teru20a}. Dataset statistics are in \cref{sec:appendix-dataset}, \cref{tab:dataset_stats}. Dataset density, defined as $2|T|/|E|$~\cite{DBLP:conf/emnlp/PujaraAG17}, is the lowest for \yago{} train (3.67) and increases across FB237 variants (i.e., from 5.33 to 9.80). This pattern also holds for the inference graphs (density ranging from 3.54 for \yago{} to 5.92 for FB237-V3), allowing us to analyze the impact of type information under varying graph sparsity. For \yago, we use the original splits with the provided ontology graph and type links; test entities are unseen during training, while relations are shared. 
Each FB237 variant contains disjoint train and inductive test graphs with distinct entities but shared relations. 
For each FB237 variant, we train on its designated training set and evaluate using its corresponding "ind" (inductive) set as the inference graph, with testing performed on its test set.
For \yago, when evaluating a specific target triple, the inference graph includes all other test triples (excluding the target itself), following \citet{zhou-etal-2023-inductive}. Since FB237 lacks concept annotations, we build ontology graphs and type links for all variants using Freebase-Wikidata mappings (see \cref{sec:appendix-dataset}).

\subsection{Metrics} 
\label{sec:experiments-metrics}
We evaluate models using Mean Reciprocal Rank (MRR) and Hits@$K$ for $K \in \{1, 10\}$, averaging over 5 evaluation runs. Following standard protocol~\cite{pmlr-v119-teru20a}, each positive test triple is ranked against 50 negative triples generated by randomly corrupting either its head or tail entity.

Tie resolution markedly affects these metrics.
While methods like random tie-breaking~\cite{DBLP:journals/tkdd/RossiBFMM21}—which randomly assign ranks among tied entities—are prevalent, they can lead to an overestimation of true model performance. This issue is particularly evident in sparse settings where limited structural or type information leads to frequent ties, an issue amplified by the candidate pool of 50.
To address these concerns and provide a more stringent and reliable evaluation, we adopt a
\textbf{strict tie-breaking strategy}. This approach assigns the positive triple the highest (i.e., worst-case/pessimistic) rank when its score is identical to one or more negative triples.

\begin{table}[t]
\centering
\small
\setlength{\tabcolsep}{12pt}
\rowcolors{2}{gray!15}{white}
\begin{tabular}{cll}
\toprule
    \textbf{ID} & \textbf{Aspect} & \textbf{Template} \\
    \midrule
    $p_1$ & type & Paris is a type of \underline{\hspace{10pt}} \\
    $p_2$ & geographic & Paris is located in \underline{\hspace{10pt}} \\
    $p_3$ & membership & Paris is member of \underline{\hspace{10pt}} \\
    $p_4$ & equivalence & Paris is equivalent to \underline{\hspace{10pt}} \\
    $p_5$ & difference & Paris is different from \underline{\hspace{10pt}} \\
    $p_6$ & similarity & Paris is similar to \underline{\hspace{10pt}} \\
    \bottomrule
\end{tabular}
\caption{Assertion prompts ($p_1$-$p_6$) used in the semantic enrichment step (\cref{sec:approach}). These templates, with a placeholder for the entity, are fed to the Pre-trained Language Model to elicit representations capturing different semantic aspects (type, geographic context, membership, equivalence, difference, similarity) of the entity.}

\label{tab:assertion-prompts}
\end{table}

\subsection{Models}

\begin{table*}[ht]
\centering
\resizebox{\textwidth}{!}{
\rowcolors{7}{white}{gray!15}
\begin{tabular}{lcccccccccccc}
\toprule
\multirow{2}{*}{\textbf{Inductive LP Model}}          &  \multicolumn{3}{c}{\textbf{FB237-V1}} & \multicolumn{3}{c}{\textbf{FB237-V2}} & \multicolumn{3}{c}{\textbf{FB237-V3}} & \multicolumn{3}{c}{\textbf{YAGO21K-610}} \\ 
\cmidrule(lr){2-4} \cmidrule(lr){5-7} \cmidrule(lr){8-10} \cmidrule(lr){11-13} 
             & MRR      & Hits@1   & Hits@10   & MRR    & Hits@1   & Hits@10   & MRR    & Hits@1   & Hits@10   & MRR    & Hits@1   & Hits@10     \\ 
\midrule
\grail \cite{pmlr-v119-teru20a}         &  .456      &  34.97    &  64.44
&    \textbf{.618}    &  \textbf{50.46}      &  \underline{82.70}     &   .609  &     \underline{49.94}   &   82.26  & \underline{.661} &  \textbf{62.76}  &  68.68 \\
\citet{zhou-etal-2023-inductive}   &    .398    &   27.85     & 64.55      &  .576    &  44.69   &    82.45   &   .554   &  41.85  &   81.42 & \textbf{.673}  & \underline{60.36}  & \underline{76.56}  \\
\tyler-RoBERTa-L (2025)     &   \underline{.470}     &  \underline{35.66}      &  \underline{69.95}      &   .602  &   47.51  &  \textbf{83.28}    & \textbf{.630} &  \textbf{50.60} & \textbf{86.72}   & .660   &  58.26  &  \textbf{79.68}          \\
\tyler-\llamathreeb (2025)      &     \textbf{.481}     &   \textbf{36.88}   &   \textbf{70.63}   &       \underline{.610}   & \underline{49.37}  &   82.01 &   \underline{.620} & \underline{49.94} &    \underline{84.46}    &  .651      &   59.70    &  69.30       \\  
\bottomrule   
\end{tabular}}
\caption{Link Prediction (LP) evaluation on multiple FB237 variants
and \yago
. 
Best and second-best scores 
are in \textbf{bold} and \underline{underlined}, respectively.
Evaluation uses the strictest tie-breaking policy (\cref{sec:experiments-metrics}), assigning the highest (worst) possible rank to the positive triple in case of ties.
}
\label{tab:results-1}
\end{table*}

To isolate the contribution of our semantic-enrichment module, we focus the comparison on methods that share a similar subgraph-reasoning backbone as Tyler. 
We evaluate \tyler against \grail~\cite{pmlr-v119-teru20a}, a type-agnostic baseline that relies solely on subgraph structure, and the ontology-enhanced method of \citet{zhou-etal-2023-inductive}, which explicitly incorporates type information via learnable embeddings and ontological constraints, even though its effectiveness is tied to the availability and quality of type annotations and ontology triples. We chose these baselines to enable a focused and meaningful comparison: (1) \grail serves as the foundational type-agnostic framework upon which many subsequent methods build~\cite{DBLP:conf/aaai/ChenHWW21, DBLP:conf/aaai/MaiZY021}; (2) the method of \citet{zhou-etal-2023-inductive} exemplifies a type-enhanced approach, with few comparable models in literature. In contrast, Tyler is designed for scenarios where explicit type information is scarce as it is able to infer implicit type semantics from PLMs.
Our semantic enrichment strategy, detailed in \cref{sec:approach}, is model-agnostic and compatible with any PLM supporting masked or causal language modeling. We use \roberta-Large~\cite{DBLP:journals/corr/abs-1907-11692} and \llamathreeb~\cite{grattafiori2024llama3herdmodels} without fine-tuning, aggregating representations from six manually crafted assertion prompts (\cref{tab:assertion-prompts}). To further evaluate the robustness of \tyler with respect to prompt selection, we report the results on FB15K-237-V1 for both \roberta and \llamathreeb across varying numbers of templates (\cref{tab:results-prompt-ablation}). The prompt aggregation function $\text{AGG}_{p}(\cdot)$ was set to SUM, as it yielded the most consistent results in our experiments (\cref{tab:results-aggregations}). Moreover, SUM offers better scalability by producing fixed-size outputs regardless of the number of prompts and may also help regularize prompt-specific noise.

\section{Results}
\label{sec:results}
Experiments aim to answer three core questions:
\begin{enumerate}[label=\textbf{RQ\arabic*.}, left=0pt, itemsep=0pt]
\item Does explicit type information improve subgraph-based inductive link prediction?
\item Can PLMs enhance node representations for subgraph-based inductive link prediction?
\item Can PLMs mitigate type and structural sparsity challenges in inductive link prediction?
\end{enumerate}

\subsection{Type Information in Subgraph-Based Inductive Link Prediction (RQ1)}

\begin{table}[t]
\centering
\resizebox{\columnwidth}{!}{
\rowcolors{2}{gray!15}{white}
\begin{tabular}{llccc}
\toprule
\textbf{PLM} & \textbf{Aggregation} & \textbf{MRR} & \textbf{Hits@1} & \textbf{Hits@10} \\
\midrule
\tyler-\robertal & TYPE-ONLY & .442 & 32.93 & 66.49 \\
\tyler-\robertal & SUM & \textbf{.470} & \textbf{35.66} & \textbf{69.95} \\
\tyler-\robertal & MEAN & \underline{.468} & \underline{35.22} & \underline{68.05} \\
\tyler-\robertal & CONCAT & .455 & 34.10 &  67.76 \\
\midrule
\tyler-\llamathreeb  & TYPE-ONLY & \underline{.477} & \textbf{37.12} & 68.29 \\
\tyler-\llamathreeb  & SUM & \textbf{.481} & \underline{36.88} & \underline{70.63} \\
\tyler-\llamathreeb & MEAN & .465 & 35.51 & 68.73 \\
\tyler-\llamathreeb & CONCAT & .474 & 35.95 &   \textbf{71.12} \\
\bottomrule
\end{tabular}
}
\caption{Ablation study on the FB237-V1 dataset, evaluating the impact of different Pre-trained Language Models and aggregation functions (Equation \ref{eq:prompt-agg}) for semantic embeddings within \tyler. 'TYPE-ONLY' uses only the representation from the $p_1$ prompt (\cref{tab:assertion-prompts}).  }
\label{tab:results-aggregations}
\end{table}
\begin{table}[ht]
\centering
\resizebox{\columnwidth}{!}{
\begin{tabular}{lcccc}
\toprule
\textbf{PLM} & \textbf{MRR} & \textbf{Hits@1} & \textbf{Hits@10}  \\
\midrule
\llamathreeb (No GNN) & .264 & 13.21 & 55.06  \\
\bottomrule
\end{tabular}}
\caption{Performance of \llamathreeb when directly evaluating the likelihood of verbalized triples on the \yago dataset by scoring them using the (negated) model perplexity.} 
\label{tab:llama3-direct-results}
\end{table}

\begin{table}[t]
\centering
\resizebox{\columnwidth}{!}{
\rowcolors{2}{gray!15}{white}
\begin{tabular}{llccc}
\toprule
\textbf{PLM} & \textbf{Template Choice ($p_i$)} & \textbf{MRR} & \textbf{Hits@1} & \textbf{Hits@10} \\
\midrule
\tyler-\robertal & 1 (TYPE-ONLY) & .442 & 32.93 & 66.49 \\
\tyler-\robertal & 1-2  & .459 & 34.97 & 68.00 \\
\tyler-\robertal & 1-3  & .451 & 34.87 & 65.32 \\
\tyler-\robertal & 1-4 & \textbf{.472} & \textbf{36.00} & 68.58 \\
\tyler-\robertal & 1-5 & .468 & 35.32 & \underline{69.46} \\
\tyler-\robertal & \textbf{1-6 (ALL)} & \underline{.470} & \underline{35.66} &  \textbf{69.95} \\
\midrule
\tyler-\llamathreeb  & 1 (TYPE-ONLY) & \underline{.477} & \underline{37.12} & 68.29 \\
\tyler-\llamathreeb  & 1-2 & .458 & 35.07 &69.36 \\
\tyler-\llamathreeb & 1-3 & .470 & 35.95 & 69.46 \\
\tyler-\llamathreeb & 1-4 & .467 & 35.22 &  \underline{70.19} \\
\tyler-\llamathreeb & 1-5 & .471 & \textbf{37.36} & 65.36 \\
\tyler-\llamathreeb & \textbf{1-6 (ALL)} & \textbf{.481} & 36.88 &   \textbf{70.63} \\
\bottomrule
\end{tabular}
}

\caption{Ablation study on the FB237-V1 dataset, evaluating the impact of chosen number of prompts (Equation \ref{eq:prompt-agg}) for semantic embeddings within \tyler. For all configurations, the aggregation function was set to SUM.}
\label{tab:results-prompt-ablation}
\end{table}

Table~\ref{tab:results-1} presents link prediction results across various models and datasets, emphasizing the role of type information in inductive link prediction. \grail, which operates without type information, performs competitively overall. It achieves strong results in both MRR and Hits@10, and obtains the highest Hits@1 on the sparse YAGO21K-610 dataset. This suggests its ability to rank correct entities precisely in low-density settings without relying on type cues.
When explicit type information is incorporated, as in~\citet{zhou-etal-2023-inductive}, performance patterns shift: while Hits@10 often remain competitive—or even surpass \grail on sparse datasets like \yago—Hits@1 consistently decline. This indicates that explicit types may mitigate sparsity by providing useful semantic signals, but also introduce complexity that reduces precision in top-ranked predictions. As dataset density increases, the performance gap between \grail and type-informed models narrows, and in some cases, \grail even outperforms the latter. This trend suggests that \textbf{explicit type information becomes less helpful—and potentially detrimental—in denser graphs}, where structural cues are already sufficient. 
In contrast, implicit type information, as leveraged by \tyler, generally leads to more robust and consistent improvements. While not always achieving the best Hits@1, models using implicit types (\tyler variants) rank first or second across most datasets for both Hits@10 and Hits@1.
These models show particular strength on sparse datasets, such as YAGO21K-610, where the gap in Hits@10 is most pronounced. This suggests that \textbf{implicit typing is more robust against topological variations in the data}, exhibiting a higher generalization potential.

\textit{Addressing \textbf{RQ1}, the benefit of explicit type information is dataset-dependent. It aids relational inference in sparse graphs lacking rich topology, but can add detrimental complexity in denser graphs. Implicit type signals, however, consistently enhance inference, mitigating structural sparsity.}


\subsection{Usefulness of PLM Representations for Implicit Type Signal (RQ2)}

The results in Table \ref{tab:results-1} provide compelling evidence that PLMs can significantly enhance node representations in subgraph-based link prediction. However, \textbf{the impact of PLMs is not uniform across all datasets}. For example, on FB237-V1 and FB237-V3, RoBERTa-L and \llamathreeb models exhibit competitive performance, especially in terms of Hits@1 and MRR, suggesting that PLMs provide a strong inductive bias for relational reasoning. In contrast, models without PLMs, like \grail, show lower performance on these datasets, particularly regarding the Hits@10 metric. This highlights the ability of PLMs to generalize and make more accurate predictions in larger, more complex graphs, where non-PLM models may struggle.
Table \ref{tab:results-aggregations} shows the impact of different aggregation strategies on the FB237-V1 dataset, with SUM showing the most consistent results. For example, \llamathreeb with SUM 
outperforms \grail across all metrics. In addition, we compare the results of different aggregation strategies with the scenario where only the "type" prompt is considered (i.e., $p_1$ in \cref{tab:assertion-prompts}), showing consistent improvements.

To comprehensively evaluate the utility of PLMs in link prediction, we further explore the potential of using LLMs directly. Specifically, we verbalize all evaluation triples (including both positive and negative candidates) in the form $label(h) \oplus label(r) \oplus label(t)$. We then employ \llamathreeb to score these verbalized triples based on their (negated) sentence perplexity and report the resulting rankings in \cref{tab:llama3-direct-results}. We exclude the FB15K-237 variants from this evaluation due to the lack of clear and consistent relational labels, which makes verbalization unreliable. The performance on \yago, which is substantially lower than that achieved by GNN-based approaches, highlights the critical importance of incorporating neighborhood structural information for this task.


\textit{Regarding \textbf{RQ2}, PLMs effectively enhance node representations for subgraph-based inductive link prediction. By providing richer semantic features, models like RoBERTa-L and \llamathreeb improve relational inference. Aggregating diverse PLM-derived semantic embeddings (i.e., from different prompts) boosts representation expressiveness.}


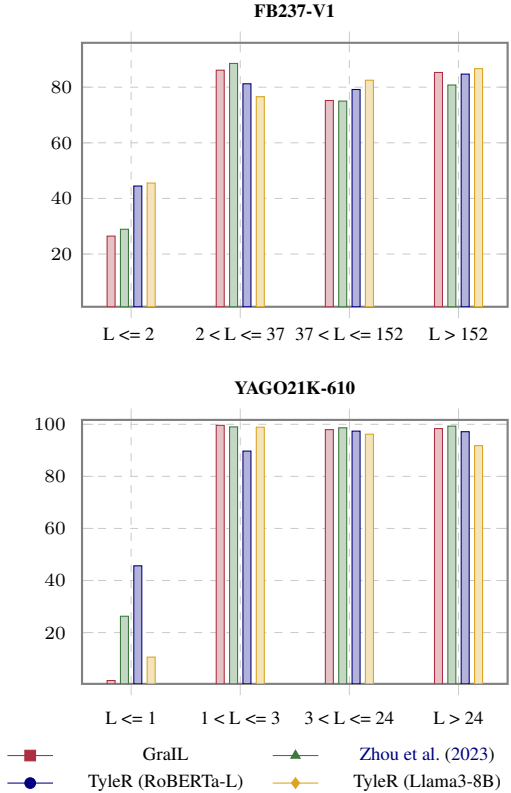
\begin{figure}[t]
\centering
\begin{tikzpicture}

\node (topplot) at (0, 0) {}; 

\begin{axis}[
    at={(topplot)},
    anchor=south,
    ybar,
    bar width=3pt,
    width=0.35\textwidth,
    height=3.5cm,
    ylabel={\empty},
    symbolic x coords={zero, one, many1, many2},
    xtick=data,
    xticklabels={L <= 2 , 2 < L <= 37, 37 < L <= 152, L > 152},
    xticklabel style={yshift=-0.0ex, align=center, font=\scriptsize},
    yticklabel style={xshift=-0.0ex, font=\scriptsize},
    ymin=12, ymax=85,
    grid=major,
    grid style={dashed, gray!40},
    tick label style={font=\scriptsize},
    label style={font=\scriptsize},
    title={\textbf{FB237-V1}},
    title style={font=\scriptsize},
    enlargelimits=0.15,
    axis lines=box,
    axis line style={gray, thick},
    tick style={gray!40},
    scale only axis,
    outer sep=0pt
]

\addplot+[color=crimson, fill=crimson!30] coordinates {(zero, 26.46) (one, 86.10) (many1, 75.21) (many2, 85.29)};
\addplot+[color=forestgreen, fill=forestgreen!30] coordinates {(zero, 28.92) (one, 88.54) (many1, 75.00) (many2, 80.78)};
\addplot+[color=darkblue, fill=darkblue!30] coordinates {(zero, 44.46) (one, 81.22) (many1, 79.17) (many2, 84.70)};
\addplot+[color=goldenrod, fill=goldenrod!30] coordinates {(zero, 45.54) (one, 76.58) (many1, 82.50) (many2, 86.67)};

\end{axis}

\begin{axis}[
    at={(topplot)}, 
    anchor=north, 
    yshift=-1.5cm, 
    ybar,
    bar width=3pt,
    width=0.35\textwidth,
    height=3.5cm,
    ylabel={\empty},
    symbolic x coords={zero, one, many1, many2},
    xtick=data,
    xticklabels={L <= 1, 1 < L <= 3, 3 < L <= 24, L > 24},
    xticklabel style={yshift=-0.5ex, align=center, font=\scriptsize},
    yticklabel style={xshift=-0.5ex, font=\scriptsize},
    ymin=12, ymax=90,
    grid=major,
    grid style={dashed, gray!40},
    tick label style={font=\scriptsize},
    label style={font=\scriptsize},
    title={\textbf{\yago}},
    title style={font=\scriptsize},
    enlargelimits=0.15,
    axis lines=box,
    axis line style={gray, thick},
    tick style={gray!40},
    scale only axis,
    outer sep=0pt
]

\addplot+[color=crimson, fill=crimson!30] coordinates {(zero, 1.58) (one, 99.61) (many1, 97.96) (many2, 98.36)};
\addplot+[color=forestgreen, fill=forestgreen!30] coordinates {(zero, 26.28) (one, 99.02) (many1, 98.67) (many2, 99.31)};
\addplot+[color=darkblue, fill=darkblue!30] coordinates {(zero, 45.65) (one, 89.72) (many1, 97.39) (many2, 97.17)};
\addplot+[color=goldenrod, fill=goldenrod!30] coordinates {(zero, 10.58) (one, 98.90) (many1, 96.21) (many2, 91.82)};

\end{axis}

\end{tikzpicture}

\begin{minipage}{\linewidth} 
\centering
\begin{tikzpicture}
\begin{axis}[
    hide axis,
    xmin=0, xmax=1,
    ymin=0, ymax=1,
    legend columns=2, 
    legend style={
        draw=none,
        font=\scriptsize,
        column sep=1em,
        /tikz/every even column/.append style={column sep=0.8em}
    }
]
\addlegendimage{crimson, mark=square*}
\addlegendentry{\grail}
\addlegendimage{forestgreen, mark=triangle*}
\addlegendentry{\citet{zhou-etal-2023-inductive}}
\addlegendimage{darkblue, mark=*}
\addlegendentry{\tyler (\robertal)}
\addlegendimage{goldenrod, mark=diamond*}
\addlegendentry{\tyler (\llamathreeb)}
\end{axis}
\end{tikzpicture}
\end{minipage}

\caption{Link Prediction (Hits@10) evaluation 
under \textbf{varying structural sparsity conditions} (i.e., the number of edges L in the enclosing subgraph of the target triple, including the target triple) on FB237-V1 (top) and \yago (bottom). 
}
\label{fig:hits10_stacked}
\end{figure}

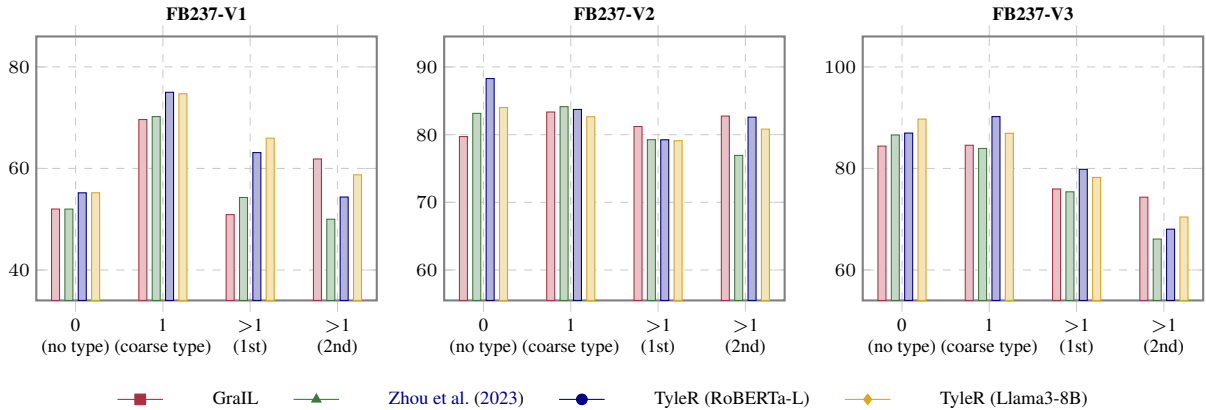
\begin{figure*}[t]
\centering
\begin{tikzpicture}

\node[anchor=south west, inner sep=0] (A) at (0,0) {
\begin{tikzpicture}
\begin{axis}[
    ybar,
    bar width=3pt,
    width=0.28\textwidth,
    height=3.5cm,
    ylabel={\empty},
    symbolic x coords={zero, one, many1, many2},
    xtick=data,
    xticklabels={0\\(no type), 1\\(coarse type), $>$1\\(1st), $>$1\\(2nd)},
    xticklabel style={yshift=-0.5ex, align=center, font=\scriptsize},
    yticklabel style={xshift=-0.5ex, font=\scriptsize},
    ymin=40, ymax=80,
    grid=major,
    grid style={dashed, gray!40},
    tick label style={font=\scriptsize},
    label style={font=\scriptsize},
    title={\textbf{FB237-V1}},
    title style={font=\scriptsize},
    enlargelimits=0.15,
    axis lines=box,
    axis line style={gray, thick},
    tick style={gray!40},
    scale only axis,
    outer sep=0pt,
    legend style={font=\tiny, at={(0.5,-0.3)}, anchor=north, legend columns=2}
]

\addplot+[color=crimson, fill=crimson!30] coordinates {(zero, 52.00) (one, 69.64) (many1, 50.91) (many2, 61.87)};
\addplot+[color=forestgreen, fill=forestgreen!30] coordinates {(zero, 52.00) (one, 70.22) (many1, 54.28) (many2, 50.00)};
\addplot+[color=darkblue, fill=darkblue!30] coordinates {(zero, 55.20) (one, 75.00) (many1, 63.12) (many2, 54.37)};
\addplot+[color=goldenrod, fill=goldenrod!30] coordinates {(zero, 55.20) (one, 74.71) (many1, 65.97) (many2, 58.75)};

\end{axis}
\end{tikzpicture}
};

\node[anchor=south west, inner sep=0] (B) at ([xshift=13pt]A.south east) {
\begin{tikzpicture}
\begin{axis}[
    ybar,
    bar width=3pt,
    width=0.28\textwidth,
    height=3.5cm,
    ylabel={\empty},
    symbolic x coords={zero, one, many1, many2},
    xtick=data,
    xticklabels={0\\(no type), 1\\(coarse type), $>$1\\(1st), $>$1\\(2nd)},
    xticklabel style={yshift=-0.5ex, align=center, font=\scriptsize},
    yticklabel style={xshift=-0.5ex, font=\scriptsize},
    ymin=60, ymax=90,
    grid=major,
    grid style={dashed, gray!40},
    tick label style={font=\scriptsize},
    label style={font=\scriptsize},
    title={\textbf{FB237-V2}},
    title style={font=\scriptsize},
    enlargelimits=0.15,
    axis lines=box,
    axis line style={gray, thick},
    tick style={gray!40},
    scale only axis,
    outer sep=0pt,
    legend style={font=\tiny, at={(0.5,-0.3)}, anchor=north, legend columns=2}
]
\addplot+[color=crimson, fill=crimson!30] coordinates {(zero, 79.71) (one, 83.34) (many1, 81.20) (many2, 82.74)};
\addplot+[color=forestgreen, fill=forestgreen!30] coordinates {(zero, 83.14) (one, 84.13) (many1, 79.25) (many2, 76.93)};
\addplot+[color=darkblue, fill=darkblue!30] coordinates {(zero, 88.28) (one, 83.72) (many1, 79.24) (many2, 82.58)};
\addplot+[color=goldenrod, fill=goldenrod!30] coordinates {(zero, 84.00) (one, 82.64) (many1, 79.10) (many2, 80.81)};

\end{axis}
\end{tikzpicture}
};

\node[anchor=south west, inner sep=0] (C) at ([xshift=13pt]B.south east) {
\begin{tikzpicture}
\begin{axis}[
    ybar,
    bar width=3pt,
    width=0.28\textwidth,
    height=3.5cm,
    ylabel={\empty},
    symbolic x coords={zero, one, many1, many2},
    xtick=data,
    xticklabels={0\\(no type), 1\\(coarse type), $>$1\\(1st), $>$1\\(2nd)},
    xticklabel style={yshift=-0.5ex, align=center, font=\scriptsize},
    yticklabel style={xshift=-0.5ex, font=\scriptsize},
    ymin=60, ymax=100,
    grid=major,
    grid style={dashed, gray!40},
    tick label style={font=\scriptsize},
    label style={font=\scriptsize},
    title={\textbf{FB237-V3}},
    title style={font=\scriptsize},
    enlargelimits=0.15,
    axis lines=box,
    axis line style={gray, thick},
    tick style={gray!40},
    scale only axis,
    outer sep=0pt
]

\addplot+[color=crimson, fill=crimson!30] coordinates {(zero, 84.40) (one, 84.57) (many1, 75.96) (many2, 74.34)};
\addplot+[color=forestgreen, fill=forestgreen!30] coordinates {(zero, 86.60) (one, 83.94) (many1, 75.40) (many2, 66.09)};
\addplot+[color=darkblue, fill=darkblue!30] coordinates {(zero, 86.97) (one, 90.21) (many1, 79.83) (many2, 68.04)};
\addplot+[color=goldenrod, fill=goldenrod!30] coordinates {(zero, 89.72) (one, 86.93) (many1, 78.24) (many2, 70.43)};

\end{axis}
\end{tikzpicture}
};
\end{tikzpicture}

\vspace{0.5em}

\begin{minipage}{\textwidth}
\centering
\begin{tikzpicture}
\begin{axis}[
    hide axis,
    xmin=0, xmax=1,
    ymin=0, ymax=1,
    legend columns=4,
    legend style={
        draw=none,
        font=\scriptsize,
        column sep=1.5em,
        /tikz/every even column/.append style={column sep=1em}
    }
]
\addlegendimage{crimson, mark=square*}
\addlegendentry{\grail}
\addlegendimage{forestgreen, mark=triangle*}
\addlegendentry{\citet{zhou-etal-2023-inductive}}
\addlegendimage{darkblue, mark=*}
\addlegendentry{\tyler (\robertal)}
\addlegendimage{goldenrod, mark=diamond*}
\addlegendentry{\tyler (\llamathreeb)}
\end{axis}
\end{tikzpicture}
\end{minipage}

\caption{Hits@10 performance across four \textbf{type sparsity} groups for three FB237 variants, computed according to the number of explicit types linked to each entity (details in \cref{sec:rq3_sparsity}).
The groups, from left to right, represent scenarios with an increasing number of explicit types associated with the known entity.
}

\label{fig:hits10_type_sparsity_fb}

\end{figure*}

\subsection{Effect on Type and Structural Sparsity (RQ3)}
\label{sec:rq3_sparsity}

To investigate the effect of our approach on \textit{type sparsity}, we categorize entities into four groups based on the number of explicit type annotations they possess. Group 0 consists of entities with \emph{no explicit type}. Group 1 includes entities with \emph{exactly one} explicit type. The remaining entities (those with \emph{more than one} type) are split into two additional groups based on the median number of types in this subset: the lower $50\%$ form the group labeled ``$>$1 (1st)'', and the upper $50\%$ form ``$>$1 (2nd)''. For each test triple, we determine the group membership of the \textit{known} entity (irrespective of the type information available for the candidate entities) and report the model performance in \cref{fig:hits10_type_sparsity_fb}. We report the average Hits@10 across the three FB237 variants for each group.
Group 0 represents the most type-sparse setting with entities lacking any explicit type; in this scenario, both variants of \tyler consistently outperform the typeless baseline \grail across all dataset variants. This reinforces the intuition that  \textbf{type signals derived from PLMs can enhance inference capabilities in sparse settings}. For instance, on FB237-V1, \tyler (\robertal) yields a 6.15\% relative improvement over the non-PLM baselines, while on FB237-V2, it achieves an even greater gain of 10.75\% over \grail. Interestingly, in the case of entities with multiple types, \tyler continues to outperform the explicit-type-based method of \citet{zhou-etal-2023-inductive}. This suggests that while explicit type information is useful, its effectiveness may diminish when type annotations are noisy or overly numerous, highlighting the need for better strategies to aggregate multiple type signals.

We further analyze the role of \textit{implicit type information} in addressing \textit{structural sparsity}. For this analysis, we consider the two datasets with the lowest graph density: \yago and FB237-V1. Evaluation triples are grouped into four bins based on the number of edges in their enclosing subgraphs, using percentiles to capture varying levels of sparsity. \cref{fig:hits10_stacked} presents Hits@10 across these structural sparsity conditions. The results indicate that PLM-based approaches, particularly those using \robertal, demonstrate strong performance in extremely sparse subgraphs. For example, \robertal performs best in scenarios with one or fewer edges (for \yago) and more than 152 edges (for FB237-V1), demonstrating its robustness at both ends of the sparsity spectrum. However, in \emph{moderate sparsity} settings (e.g., $2 < L \leq 37$ in FB237-V1), models such as \grail and \citet{zhou-etal-2023-inductive} perform comparably or better, due to their reliance on structural patterns that are still informative in such contexts.

\textit{Answering \textbf{RQ3}, PLM-based approaches such as \tyler address both type and structural sparsity. They consistently outperform baselines in scenarios with minimal explicit type information or sparse subgraph structures by inferring meaningful semantics from PLMs. Although challenges persist with moderate sparsity and noisy types, PLMs show significant potential.}


\section{Related Work}
\label{sec:related}


\paragraph{Inductive Link Prediction.}
Inductive Link Prediction (ILP) in Knowledge Graphs (KGs) aims to infer missing links that involve entities unseen during training, thereby enabling models to generalize to evolving KGs. Unlike traditional embedding-based models~\cite{DBLP:conf/aaai/LinLSLZ15, DBLP:conf/nips/BordesUGWY13, DBLP:conf/aaai/WangZFC14}, inductive methods explicitly handle unseen entities. Early approaches relied on rule-based reasoning~\cite{DBLP:conf/nips/YangYC17, DBLP:conf/semweb/MeilickeFWRGS18}, but graph neural networks (GNNs) soon became dominant~\cite{DBLP:conf/nips/HamiltonYL17}, with \grail~\cite{pmlr-v119-teru20a} leveraging enclosing subgraph structures for relational inference. Extensions include CoMPILE~\cite{DBLP:conf/aaai/MaiZY021}, emphasizing relational directionality, and TACT~\cite{DBLP:conf/aaai/ChenHWW21}, introducing relation-level reasoning. \citet{zhou-etal-2023-inductive} incorporate ontological data, but assume complete type information, an assumption that seldom holds in real-world KGs.

\paragraph{Entity Representation with Language Models.}
Pre-trained language models (PLMs) capture factual and relational knowledge from large corpora~\cite{petroni-etal-2019-language, DBLP:conf/nips/BrownMRSKDNSSAA20}, encoding rich entity semantics~\cite{DBLP:journals/www/ZhuWCQOYDCZ24} and retrieving factual information via prompting~\cite{DBLP:conf/emnlp/WeiH0K23}. This makes PLMs well-suited for link prediction because they can enrich entity representations. For example, KGBERT~\cite{DBLP:journals/corr/KGBERT} verbalizes triples as text and fine-tunes BERT to classify their plausibility. Subsequent methods~\cite{DBLP:conf/emnlp/ZhangLZ00H20, DBLP:conf/www/DazaCG21, DBLP:journals/tacl/WangGZZLLT21} integrate entity descriptions into KG completion to induce embeddings for new entities via PLMs. BERTRL \cite{DBLP:conf/aaai/ZhaCY22} exemplifies this trend by injecting GNN-discovered reasoning paths into a BERT-based model. A promising direction involves integrating LLMs with subgraph-based methods to reduce model queries while preserving structural reasoning. \citet{DBLP:conf/aaai/LiYXSJGLK25} propose CATS, a hybrid model that leverages latent type cues and neighbor facts to fine-tune an LLM for triple scoring, combining semantic understanding with explicit subgraph evidence. 
Unlike prior approaches that fine-tune PLMs, our method extracts semantic knowledge from a frozen PLM, and we investigate how effectively such pre-trained models enable a subgraph-reasoning module to capture the type semantics underlying each relation.

\section{Conclusion}
\label{sec:conclusion}

We present \tyler, a novel inductive link-prediction approach designed to handle incomplete or noisy type information. By leveraging pre-trained language models (PLMs), \tyler enriches node representations with implicit type signals, overcoming the limitations of methods reliant on explicit annotations. 
Experiments show that \tyler exhibits competitive performance, particularly when type data are sparse or unreliable. The results underscore the potential of PLMs for semantic enrichment, enabling robust link prediction without complete type supervision.
Future work will examine domain-specific PLMs, more embedding-aggregation strategies, and broader applications to graph-based tasks.


\section*{Limitations}
\label{sec:limitations}

Our study employs a set of predefined prompts, which, while effective for the scope of our experiments, may not represent the most informative or optimal configurations. More sophisticated strategies for adaptive prompt selection or prompt tuning could potentially enhance model performance. Exploring these approaches is left as a direction for future research. Additionally, the hyperparameters for our models were selected empirically, based on extensive experimentation and informed judgment. While this approach yielded strong results, it may not guarantee optimal configurations. A more systematic or exhaustive hyperparameter search could lead to improved outcomes. Nonetheless, the computational cost and complexity associated with such procedures, particularly given the scale and resource demands of our training setup, render them infeasible within the constraints of this study.

\section*{Acknowledgments}

This study was supported by OVS: Fashion Retail Reloaded, Lutech Digitale 4.0, Natuzzi S.p.A. del Contratto di Sviluppo Industriale ai sensi dell’art. 9 del Decreto del Ministro dello Sviluppo Economico del 09.12.2014, EPANSA (FAIR) - Enhancing Personal Assistants with Neuro-Symbolic AI and Knowledge Graphs, funded by the European Union Next- GenerationEU (NRRP – M4C2, Investment 1.3, D.R. No. 123 of 16/01/2024, PE00000013, CUP: H97G22000210007), “Patto Territoriale sistema universitario pugliese” CUP F61B23000370006- cod.id. PATTI\_TERRITORIALI\_WP1, XAI4AMELIA - Implementation of Explainable Artificial Intelligence Technologies for Advancing Interoperability and Analysis in AMELIA, funded by the European Union Next- GenerationEU (NRRP – M4C2, Investment 1.3, D.D. 341 del 15/03/2022, CUP: J33C22002910001). We acknowledge ISCRA for awarding this project access to the LEONARDO supercomputer, owned by the EuroHPC Joint Undertaking, hosted by CINECA (Italy). This work has been carried out while Giovanni Servedio was enrolled in the Italian National Doctorate on Artificial Intelligence run by Sapienza University of Rome in collaboration with Politecnico di Bari.

\bibliography{custom_full}

\clearpage
\appendix
\section*{Appendix}
\noindent This appendix provides supplementary material to support the main paper. It is organized as follows:
\begin{itemize}
\item \textbf{Dataset Details (\cref{sec:appendix-dataset})}: Provides a summary table (\cref{tab:dataset_stats}) and further information on the datasets used in our experiments. The section includes the procedure for extracting ontology graphs and entity-type links for the FB237 variants, which initially lack such annotations. It also details the methodology for splitting ontology triples into training, validation, and test sets.
\item \textbf{Hyperparameter Details (\cref{sec:appendix-hyperparametes})}: Outlines the hyperparameter settings employed for training our proposed model, \tyler, as well as the baseline models. Key parameters such as learning rates, number of hops for subgraph extraction, embedding dimensions, and early stopping criteria are specified to ensure reproducibility.
\item \textbf{Examples of Predictions (\cref{sec:appendix-examples})}: Presents a qualitative example (\cref{tab:triple_ranking_comparison}) comparing the link predictions made by \tyler and baseline models for a specific target triple, particularly in a scenario with a sparse enclosing subgraph. This section illustrates how different models rank candidate entities and highlights the impact of the strict tie-breaking strategy.
\item \textbf{Embedding Visualization (\cref{sec:appendix-visualization})}: Includes a 2D visualization of entity embeddings (obtained via PCA). This offers a qualitative insight into the learned representations and their spatial distribution for a sample set of entities (\cref{fig:emb_visualization_zhou} and \cref{fig:emb_visualization_tyler}).
\end{itemize}

\section{Dataset Details}
\label{sec:appendix-dataset}

\begin{table*}[hbt]
    \centering
    \footnotesize
    \renewcommand{\arraystretch}{1.15}
    \setlength{\tabcolsep}{4.5pt}
    \begin{tabular}{llcccccccc}
        \toprule
        \textbf{Dataset} & \textbf{Split} & \textbf{Entities} & \textbf{Relations} & \textbf{Triples} & \textbf{Types} & \textbf{Meta Rel.} & \textbf{Onto. Triples} & \textbf{Type Links}  & \textbf{Text Labels}\\
        \midrule
        & train & 1594 & 180 & 4245 & 458 & 29 & 680 & 2163 & 1516 \\
         \rowcolor{gray!15} 
        & valid & 567 & 103 & 489 & 124 & 15 & 86 & 756  & 539\\
               \multirow{-3}{*}{\textbf{fb237\_v1}} & test  & 550  & 102  & 492 & 113  & 13  & 85  & 764  & 517\\
        \hline
        \rowcolor{gray!15} 
        \multirow{3}{*}{\textbf{fb237\_v1\_ind}} & train & 1093 & 142 & 1993 & 458 & 29 & 680 & 1525 & 1041 \\
        & valid & 287 & 66 & 206 & 124 & 15 & 86 & 406  & 275\\
        \rowcolor{gray!15} 
        & test  & 301  & 68  & 205 & 113  & 13  & 85  & 434  & 289\\
        \hline
        & train & 2608 & 200 & 9739 & 575 & 33 & 865 & 3586 & 2489 \\
        \rowcolor{gray!15} 
        & valid & 1139 & 143 & 1166 & 160 & 15 & 109 & 1511  & 1083\\
        \multirow{-3}{*}{\textbf{fb237\_v2}} & test  & 1142  & 140  & 1180 & 153  & 16  & 108  & 1515  & 1094\\
        \hline
        \rowcolor{gray!15} 
        \multirow{3}{*}{\textbf{fb237\_v2\_ind}} & train & 1660 & 172 & 4145 & 575 & 33 & 865 & 2257 & 1561 \\
        & valid & 548 & 92 & 469 & 160 & 15 & 109 & 757  & 516\\
         \rowcolor{gray!15} 
        & test  & 562  & 107  & 478 & 153  & 16  & 108  & 745  & 524\\
        \hline
         & train & 3668 & 215 & 17986 & 732 & 31 & 1060 & 5114 & 3484 \\
         \rowcolor{gray!15} 
        & valid & 1882 & 183 & 2194 & 196 & 17 & 133 & 2575  & 1787\\
        \multirow{-3}{*}{\textbf{fb237\_v3}}
        & test  & 1871  & 179  & 2214 & 192  & 16  & 133  & 2520  & 1773\\
        \hline
        \rowcolor{gray!15} 
        \multirow{3}{*}{\textbf{fb237\_v3\_ind}} & train & 2501 & 183 & 7406 & 732 & 31 & 1060 & 3426 & 2379 \\
        & valid & 973 & 120 & 866 & 196 & 17 & 133 & 1275  & 920\\
        \rowcolor{gray!15} 
        & test  & 981  & 128  & 865 & 192  & 16  & 133  & 1290  & 924\\
        \hline
            & train & 16357 &  30 & 30000 & 610 & 24 & 1983 & 4861 & 16357\\
             \rowcolor{gray!15}
            & valid &  4388 &  21 &  3000 & 166 & 14 &  248 & 1783 & 4388\\
                \multirow{-3}{*}{\textbf{YAGO21K-610}} & test  &  3938 &  25 &  6970 & 159 & 13 &  248 & 1898 & 3938\\
        \bottomrule
    \end{tabular}
    \captionof{table}{Statistics of the datasets used in our experiments. The \yago~\cite{zhou-etal-2023-inductive} dataset includes ontology triples and entity-type links, while the FB237 dataset variants~\cite{pmlr-v119-teru20a} are further processed to extract ontology triples, type links and textual labels.}
    \label{tab:dataset_stats}
\end{table*}

\begin{table*}[t]
\centering
\footnotesize
\setlength{\tabcolsep}{14pt}
\rowcolors{3}{white}{gray!15}
\begin{tabular}{r l r}
\toprule
\multicolumn{1}{c}{\textbf{Rank}} & \multicolumn{1}{c}{\textbf{Triple}} & \multicolumn{1}{c}{\textbf{Score}} \\
\midrule
\multicolumn{3}{c}{\textbf{\tyler (\robertal)}} \\
\midrule
1 & Christos Kagiouzis $\rightarrow$ \texttt{isAffiliatedTo} $\rightarrow$ Southern United FC & -1.951 \\
\rowcolor{yellow!50}
\textbf{2} & \textbf{Christos Kagiouzis $\rightarrow$ \texttt{isAffiliatedTo} $\rightarrow$ Kastoria F.C. \quad (\textit{gold})} & \textbf{-1.953} \\
3 & Christos Kagiouzis $\rightarrow$ \texttt{isAffiliatedTo} $\rightarrow$ Deltras F.C. & -1.985 \\
4 & Christos Kagiouzis $\rightarrow$ \texttt{isAffiliatedTo} $\rightarrow$ Yunnan Hongta F.C. & -1.998 \\
5 & Christos Kagiouzis $\rightarrow$ \texttt{isAffiliatedTo} $\rightarrow$ Chainat Hornbill F.C. & -4.957 \\
6 & Christos Kagiouzis $\rightarrow$ \texttt{isAffiliatedTo} $\rightarrow$ Hoàng Anh Gia Lai F.C. & -5.172 \\
7 & Christos Kagiouzis $\rightarrow$ \texttt{isAffiliatedTo} $\rightarrow$ Great Britain women's Olympic football team & -5.529 \\
8 & Christos Kagiouzis $\rightarrow$ \texttt{isAffiliatedTo} $\rightarrow$ APEP F.C. & -5.604 \\
9 & Christos Kagiouzis $\rightarrow$ \texttt{isAffiliatedTo} $\rightarrow$ Basketball League Belgium & -5.698 \\
10 & Christos Kagiouzis $\rightarrow$ \texttt{isAffiliatedTo} $\rightarrow$ Baltimore Blast (1980–92) & -5.706 \\
\midrule
\midrule
\multicolumn{3}{c}{\textbf{\grail}} \rowcolors{3}{gray!15}{white}\\
\midrule

41 & Christos Kagiouzis $\rightarrow$ \texttt{isAffiliatedTo} $\rightarrow$ Darko Vukić \phantom{xxxxxxxxxxxxxxxxxxx} & -11.888 \\
\rowcolor{gray!15} 42 & Christos Kagiouzis $\rightarrow$ \texttt{isAffiliatedTo} $\rightarrow$ Connecticut Pride & -11.888 \\
43 & Christos Kagiouzis $\rightarrow$ \texttt{isAffiliatedTo} $\rightarrow$ Hoàng Anh Gia Lai F.C. & -11.888 \\
44 & Christos Kagiouzis $\rightarrow$ \texttt{isAffiliatedTo} $\rightarrow$ Southern United FC & -11.888 \\
45 & Christos Kagiouzis $\rightarrow$ \texttt{isAffiliatedTo} $\rightarrow$ Helgi Sigurðsson & -11.888 \\
46 & Christos Kagiouzis $\rightarrow$ \texttt{isAffiliatedTo} $\rightarrow$ Conor Powell & -11.888 \\
47 & Christos Kagiouzis $\rightarrow$ \texttt{isAffiliatedTo} $\rightarrow$ Samuel Cunningham (footballer) & -11.888 \\
48 & Christos Kagiouzis $\rightarrow$ \texttt{isAffiliatedTo} $\rightarrow$ Ferdinand Daučík & -11.888 \\
49 & Christos Kagiouzis $\rightarrow$ \texttt{isAffiliatedTo} $\rightarrow$ Ertan Demiri & -11.888 \\
\rowcolor{yellow!50}
\textbf{50} & \textbf{Christos Kagiouzis $\rightarrow$ \texttt{isAffiliatedTo} $\rightarrow$ Kastoria F.C. \quad (\textit{gold})} & \textbf{-11.888} \\
\midrule
\midrule
\multicolumn{3}{c}{\textbf{\citet{zhou-etal-2023-inductive}}} \rowcolors{3}{gray!15}{white}\\
\midrule
10 & Christos Kagiouzis $\rightarrow$ \texttt{isAffiliatedTo} $\rightarrow$ SC 07 Bad Neuenahr \phantom{xxxxxxxxxxx} & 4.330 \\
\rowcolor{gray!15} 11 & Christos Kagiouzis $\rightarrow$ \texttt{isAffiliatedTo} $\rightarrow$ Chicago Power & 4.330 \\
12 & Christos Kagiouzis $\rightarrow$ \texttt{isAffiliatedTo} $\rightarrow$ Baltimore Blast (1980–92) & 4.330 \\
13 & Christos Kagiouzis $\rightarrow$ \texttt{isAffiliatedTo} $\rightarrow$ Peristeri B.C. & 4.330 \\
14 & Christos Kagiouzis $\rightarrow$ \texttt{isAffiliatedTo} $\rightarrow$ Deltras F.C. & 4.330 \\
15 & Christos Kagiouzis $\rightarrow$ \texttt{isAffiliatedTo} $\rightarrow$ ADET & 4.330 \\
\rowcolor{yellow!50}
\textbf{16} & \textbf{Christos Kagiouzis $\rightarrow$ \texttt{isAffiliatedTo} $\rightarrow$ Kastoria F.C. \quad (\textit{gold})} & \textbf{4.330} \\
17 & Christos Kagiouzis $\rightarrow$ \texttt{isAffiliatedTo} $\rightarrow$ Łukasz Tumicz & -6.757 \\
18 & Christos Kagiouzis $\rightarrow$ \texttt{isAffiliatedTo} $\rightarrow$ Ertan Demiri & -6.757 \\
19 & Christos Kagiouzis $\rightarrow$ \texttt{isAffiliatedTo} $\rightarrow$ Ferdinand Daučík & -6.757 \\
\bottomrule
\end{tabular}

\caption{Example of ranking predictions on the \yago dataset for the target triple \texttt{(Christos Kagiouzis, isAffiliatedTo, Kastoria F.C.)}, when the tail is to be predicted. In this case, the target triple has no links in the associated enclosing subgraph. As discussed in \cref{sec:experiments-metrics}, ranking is done using the strict tie-breaking strategy.}
\label{tab:triple_ranking_comparison}
\end{table*}

\cref{tab:dataset_stats} provides a statistical overview of the datasets utilized in our experiments, detailing their key characteristics, including the number of entities, relations, triples, types, meta-relations, ontology triples, type links, and textual labels.

The model from \citet{zhou-etal-2023-inductive} relies on explicit entity-type pairs and an ontology graph for training. FB237 initially lacks these annotations. Therefore, we processed the FB237 variants to extract the necessary type information and construct a corresponding ontology using the following procedure.

To construct the ontology graph for our experiments we mapped all the freebase entities appearing in the dataset to their Wikidata identifier, using the publicly available Freebase-Wikidata mappings~\footnote{https://developers.google.com/freebase}.
Using the public Wikidata API \footnote{https://www.wikidata.org/w/api.php}, we then retrieved for every mapped entity its respective textual label and the values associated with its “\textit{instance of}” property, which indicates the type(s) an entity is associated to.
With the set of relevant concepts established, we constructed the schema-level ontology.
For each concept identified in the previous step, its full set of concepts was fetched from Wikidata.

A schema-level triple ⟨$\text{Concept}_1$ PropertyLabel $\text{Concept}_2$⟩ was generated and added to our ontology graph if, and only if, the target value of a Concept ($Concept_2$) was itself one of the recognized concepts. 

In the entity triples, the entities in the test set do not appear in the train
set and valid set, while the relations in both the test set and valid set are included in the train set.
We train on the train graph and test on the test graph. 
In addition, to achieve ontology training, we randomly divide the ontology triples into a train set, a valid set, and a test set using hold-out splitting in the ratio of $80\%$, $10\%$, $10\%$, respectively.

\section{Hyperparameter Details}
\label{sec:appendix-hyperparametes}

Baselines are trained using the hyperparameter settings reported in their original papers. For our model, we adopt the configuration from \citet{zhou-etal-2023-inductive} to ensure fair comparison, tuning only the learning rate, which we set empirically to 1e-3. All models are trained for 50 epochs with early stopping (patience of 100 iterations) and a batch size of 16. We adopt the Adam optimizer. For all models, the number of hops in the enclosing subgraph is 3. We set the semantic embedding dimension to 24, the layer-0 embedding dimension to 32, and the margin $\gamma$ in the loss function to 10.

\section{Examples of Predictions}
\label{sec:appendix-examples}

This section provides a qualitative example to illustrate the behavior of \tyler in comparison to baseline models, particularly in challenging scenarios characterized by extreme structural sparsity. We focus on a specific instance from the \yago dataset where the enclosing subgraph for the target triple lacks any connecting edges.

\cref{tab:triple_ranking_comparison} presents the top-ranked predictions for the target triple \texttt{(Christos Kagiouzis, isAffiliatedTo, Kastoria F.C.)}, where the task is to predict the tail entity \texttt{(Kastoria F.C.)}. 
This triple was chosen because its $3$-hop enclosing subgraph presents a worst-case scenario for structural reasoning. Specifically, the subgraph contains no path that could link the head entity (\texttt{Christos Kagiouzis}) to the correct tail entity (\texttt{Kastoria F.C.}), beside the target link.
This lack of structural information within the subgraph presents a significant challenge for models that heavily rely on graph patterns. The evaluation follows the standard protocol (\cref{sec:experiments-metrics}), where the correct tail entity is ranked against 50 randomly corrupted negative samples. Crucially, as detailed in \cref{sec:experiments-metrics}, ranking employs the strict tie-breaking strategy, assigning the worst possible rank to the positive triple in case of score ties.

\subsection{Analysis}

\textbf{\tyler (RoBERTa-L).} Despite the absence of direct structural paths in the enclosing subgraph, \tyler ranks the correct entity \texttt{(Kastoria F.C.)} 2nd. This strong performance is attributed to its ability to leverage rich semantic information derived from the PLM (RoBERTa-L). The PLM's understanding of entities and their likely affiliations, learned from vast text corpora, allows \tyler to infer plausible connections even when explicit graph structure is missing. The top-ranked entity, \texttt{(Southern United FC)}, is also a football club, indicating that \tyler correctly identifies the semantic category of plausible tail entities for the relation \texttt{(isAffiliatedTo)} with \texttt{(Christos Kagiouzis)} (likely a footballer). The scores assigned by \tyler  are relatively distinct, suggesting a higher degree of confidence in its ranking.

\noindent \textbf{GraIL.} In contrast, GraIL, which relies purely on subgraph structures for relational inference, performs poorly. It ranks the correct entity \texttt{(Kastoria F.C.)} at 50th (last among the 50 candidates considered for ranking this positive triple). The identical scores for all top 50 entities (all -11.888) indicate that GraIL cannot differentiate between the candidates due to the lack of structural cues in the enclosing subgraph. This highlights a key limitation of purely structural methods in extremely sparse settings.

\noindent \textbf{\citet{zhou-etal-2023-inductive}.} This model, which incorporates explicit type information and ontology reasoning, ranks the correct entity 16th. While this is significantly better than GraIL, it falls short of \tyler's performance. The explicit type information likely provides some signal ("\texttt{(Kastoria F.C.)} is a \textit{Club}"). However, this explicit information might be coarser-grained or less directly informative for this specific prediction compared to the nuanced semantic representations captured by \tyler. The presence of many ties in the scores (e.g., ranks 10-16 all have score 4.330) suggests that while types help narrow down possibilities, they do not offer the same fine-grained discriminative power as \tyler's PLM-based semantic enrichment in this particular sparse scenario.

\vspace{0.5em}
\textit{This example underscores the advantage of \tyler's approach, particularly its semantic enrichment stage using PLMs. By infusing node representations with implicit type-aware signals, \tyler can effectively reason about entity relationships even when the local graph structure is uninformative, thereby mitigating the challenges posed by structural sparsity.}

\section{Embedding Visualization}
\label{sec:appendix-visualization}




This section provides a qualitative analysis of entity embeddings through 2D visualization to illustrate how different models represent candidate entities in a challenging link prediction task characterized by structural sparsity. We utilize Principal Component Analysis (PCA) to project the final-layer GNN embeddings $\mathbf{h}_v^L$ of 50 candidate tail entities onto a 2D plane. The specific task visualized is predicting the missing tail entity for the triple <\texttt{Andrei Gashkin, playsFor, ?}> from the \yago dataset.
Notably, this example is chosen for its extreme structural sparsity. The enclosing subgraph constructed around the head entity \texttt{Andrei Gashkin} and the correct tail entity \texttt{FC KAMAZ Naberezhnye Chelny} is very sparse. Furthermore, for many of the 49 negative candidate entities considered alongside the correct tail, their respective enclosing subgraphs (when considered with the head \texttt{Andrei Gashkin}) also lack rich structural information, making it difficult for models relying heavily on graph patterns to make accurate distinctions.
We compare the embeddings generated by:
\begin{itemize}
\item The ontology-enhanced model from~\citet{zhou-etal-2023-inductive}, which leverages explicit type information (\cref{fig:emb_visualization_zhou}).
\item Our proposed model, \tyler (\robertal), which uses PLM-derived implicit type signals (\cref{fig:emb_visualization_tyler}).
\end{itemize}

\subsection{Analysis} 
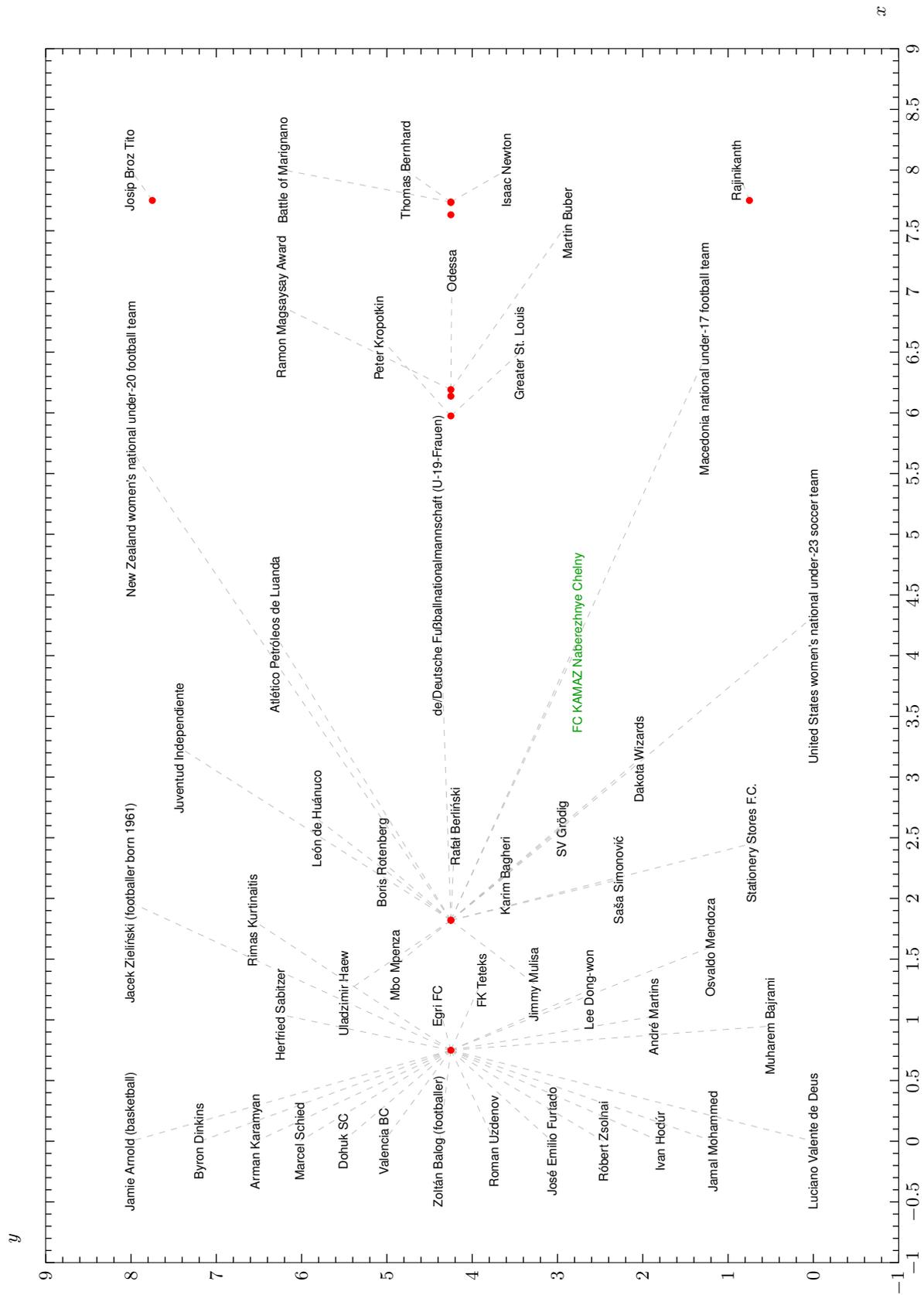
\begin{figure*}
\rotatebox{90}{%
\centering
\begin{tikzpicture}
\begin{axis}[
    width=1.4\textwidth,
    height=0.65\textheight,
    xlabel={$x$},
    ylabel={$y$},
    xmin=-1, xmax=9,
    ymin=-1, ymax=9,
    grid=none,
    grid style={line width=.1pt, draw=gray!30},
    major grid style={line width=.2pt, draw=gray!50},
    minor tick num=4,
    tick style={black, line width=0.5pt},
    axis line style={-latex},
    xlabel style={at={(axis description cs:1.02,0.02)}, anchor=west, font=\small\sffamily},
    ylabel style={rotate=-90, at={(axis description cs:0.02,1.02)}, anchor=south, font=\small\sffamily},
    ticklabel style={font=\small\sffamily},
    clip=false
]

\addplot[mark=*, mark size=1.5pt, color=red] coordinates {(1.8211982,4.2500000)};
\addplot[mark=*, mark size=1.5pt, color=red] coordinates {(0.7500000,4.2500000)};
\addplot[mark=*, mark size=1.5pt, color=red] coordinates {(1.8194728,4.2500000)};
\addplot[mark=*, mark size=1.5pt, color=red] coordinates {(7.6317695,4.2500000)};
\addplot[mark=*, mark size=1.5pt, color=red] coordinates {(6.1375026,4.2500000)};
\addplot[mark=*, mark size=1.5pt, color=red] coordinates {(7.7328039,4.2500000)};
\addplot[mark=*, mark size=1.5pt, color=red] coordinates {(7.7499953,7.7500000)};
\addplot[mark=*, mark size=1.5pt, color=red] coordinates {(6.1915157,4.2500000)};
\addplot[mark=*, mark size=1.5pt, color=red] coordinates {(7.7373680,4.2500000)};
\addplot[mark=*, mark size=1.5pt, color=red] coordinates {(5.9747529,4.2500000)};
\addplot[mark=*, mark size=1.5pt, color=red] coordinates {(7.7500000,0.7500000)};
\draw[dashed, gray!50, line width=0.3pt] (1.8211982,4.2500000) -- (4.1073667,2.7514810);
\node[font=\tiny\sffamily, align=center, inner sep=1pt, text=green!60!black] at (4.1073667,2.7514810) {FC KAMAZ Naberezhnye Chelny};
\draw[dashed, gray!50, line width=0.3pt] (1.8211982,4.2500000) -- (4.1757441,6.3144958);
\node[font=\tiny\sffamily, align=center, inner sep=1pt] at (4.1757441,6.3144958) {Atlético Petróleos de Luanda};
\draw[dashed, gray!50, line width=0.3pt] (1.8211982,4.2500000) -- (3.2389520,7.4197914);
\node[font=\tiny\sffamily, align=center, inner sep=1pt] at (3.2389520,7.4197914) {Juventud Independiente};
\draw[dashed, gray!50, line width=0.3pt] (1.8194728,4.2500000) -- (2.6580116,5.8217917);
\node[font=\tiny\sffamily, align=center, inner sep=1pt] at (2.6580116,5.8217917) {León de Huánuco};
\draw[dashed, gray!50, line width=0.3pt] (0.7500000,4.2500000) -- (1.1141214,4.3900459);
\node[font=\tiny\sffamily, align=center, inner sep=1pt] at (1.1141214,4.3900459) {Egri FC};
\draw[dashed, gray!50, line width=0.3pt] (0.7500000,4.2500000) -- (1.3248275,3.8845040);
\node[font=\tiny\sffamily, align=center, inner sep=1pt] at (1.3248275,3.8845040) {FK Teteks};
\draw[dashed, gray!50, line width=0.3pt] (1.8194728,4.2500000) -- (2.4546841,0.7098962);
\node[font=\tiny\sffamily, align=center, inner sep=1pt] at (2.4546841,0.7098962) {Stationery Stores F.C.};
\draw[dashed, gray!50, line width=0.3pt] (1.8211982,4.2500000) -- (2.5733586,2.9367714);
\node[font=\tiny\sffamily, align=center, inner sep=1pt] at (2.5733586,2.9367714) {SV Grödig};
\draw[dashed, gray!50, line width=0.3pt] (1.8211982,4.2500000) -- (3.1476112,2.0410794);
\node[font=\tiny\sffamily, align=center, inner sep=1pt] at (3.1476112,2.0410794) {Dakota Wizards};
\draw[dashed, gray!50, line width=0.3pt] (1.8194728,4.2500000) -- (2.5957532,4.1989035);
\node[font=\tiny\sffamily, align=center, inner sep=1pt] at (2.5957532,4.1989035) {Rafał Berliński};
\draw[dashed, gray!50, line width=0.3pt] (1.8194728,4.2500000) -- (2.2961264,5.0415485);
\node[font=\tiny\sffamily, align=center, inner sep=1pt] at (2.2961264,5.0415485) {Boris Rotenberg};
\draw[dashed, gray!50, line width=0.3pt] (1.8194728,4.2500000) -- (1.4540437,4.8871754);
\node[font=\tiny\sffamily, align=center, inner sep=1pt] at (1.4540437,4.8871754) {Mbo Mpenza};
\draw[dashed, gray!50, line width=0.3pt] (1.8194728,4.2500000) -- (1.2184992,5.5040264);
\node[font=\tiny\sffamily, align=center, inner sep=1pt] at (1.2184992,5.5040264) {Uladzimir Haew};
\draw[dashed, gray!50, line width=0.3pt] (0.7500000,4.2500000) -- (1.8213637,6.5726066);
\node[font=\tiny\sffamily, align=center, inner sep=1pt] at (1.8213637,6.5726066) {Rimas Kurtinaitis};
\draw[dashed, gray!50, line width=0.3pt] (0.7500000,4.2500000) -- (1.9650495,8.0000000);
\node[font=\tiny\sffamily, align=center, inner sep=1pt] at (1.9650495,8.0000000) {Jacek Zieliński (footballer born 1961)};
\draw[dashed, gray!50, line width=0.3pt] (0.7500000,4.2500000) -- (1.0437177,6.2502585);
\node[font=\tiny\sffamily, align=center, inner sep=1pt] at (1.0437177,6.2502585) {Herfried Sabitzer};
\draw[dashed, gray!50, line width=0.3pt] (0.7500000,4.2500000) -- (0.0000000,8.0000000);
\node[font=\tiny\sffamily, align=center, inner sep=1pt] at (0.0000000,8.0000000) {Jamie Arnold (basketball)};
\draw[dashed, gray!50, line width=0.3pt] (0.7500000,4.2500000) -- (0.0000000,7.1871175);
\node[font=\tiny\sffamily, align=center, inner sep=1pt] at (0.0000000,7.1871175) {Byron Dinkins};
\draw[dashed, gray!50, line width=0.3pt] (0.7500000,4.2500000) -- (0.0000000,6.5293728);
\node[font=\tiny\sffamily, align=center, inner sep=1pt] at (0.0000000,6.5293728) {Arman Karamyan};
\draw[dashed, gray!50, line width=0.3pt] (0.7500000,4.2500000) -- (0.0000000,6.0209152);
\node[font=\tiny\sffamily, align=center, inner sep=1pt] at (0.0000000,6.0209152) {Marcel Schied};
\draw[dashed, gray!50, line width=0.3pt] (0.7500000,4.2500000) -- (0.0000000,5.5118726);
\node[font=\tiny\sffamily, align=center, inner sep=1pt] at (0.0000000,5.5118726) {Dohuk SC};
\draw[dashed, gray!50, line width=0.3pt] (0.7500000,4.2500000) -- (0.0000000,5.0340364);
\node[font=\tiny\sffamily, align=center, inner sep=1pt] at (0.0000000,5.0340364) {Valencia BC};
\draw[dashed, gray!50, line width=0.3pt] (0.7500000,4.2500000) -- (0.0000000,4.3835794);
\node[font=\tiny\sffamily, align=center, inner sep=1pt] at (0.0000000,4.3835794) {Zoltán Balog (footballer)};
\draw[dashed, gray!50, line width=0.3pt] (0.7500000,4.2500000) -- (0.0000000,3.7413870);
\node[font=\tiny\sffamily, align=center, inner sep=1pt] at (0.0000000,3.7413870) {Roman Uzdenov};
\draw[dashed, gray!50, line width=0.3pt] (0.7500000,4.2500000) -- (0.0000000,3.0642495);
\node[font=\tiny\sffamily, align=center, inner sep=1pt] at (0.0000000,3.0642495) {José Emilio Furtado};
\draw[dashed, gray!50, line width=0.3pt] (0.7500000,4.2500000) -- (0.0000000,2.4593391);
\node[font=\tiny\sffamily, align=center, inner sep=1pt] at (0.0000000,2.4593391) {Róbert Zsolnai};
\draw[dashed, gray!50, line width=0.3pt] (0.7500000,4.2500000) -- (0.0000000,1.7960103);
\node[font=\tiny\sffamily, align=center, inner sep=1pt] at (0.0000000,1.7960103) {Ivan Hodúr};
\draw[dashed, gray!50, line width=0.3pt] (0.7500000,4.2500000) -- (0.0000000,1.1749858);
\node[font=\tiny\sffamily, align=center, inner sep=1pt] at (0.0000000,1.1749858) {Jamal Mohammed};
\draw[dashed, gray!50, line width=0.3pt] (0.7500000,4.2500000) -- (0.0000000,0.0000000);
\node[font=\tiny\sffamily, align=center, inner sep=1pt] at (0.0000000,0.0000000) {Luciano Valente de Deus};
\draw[dashed, gray!50, line width=0.3pt] (0.7500000,4.2500000) -- (1.0291927,1.8704889);
\node[font=\tiny\sffamily, align=center, inner sep=1pt] at (1.0291927,1.8704889) {André Martins};
\draw[dashed, gray!50, line width=0.3pt] (0.7500000,4.2500000) -- (0.9493685,0.4905116);
\node[font=\tiny\sffamily, align=center, inner sep=1pt] at (0.9493685,0.4905116) {Muharem Bajrami};
\draw[dashed, gray!50, line width=0.3pt] (0.7500000,4.2500000) -- (1.5977260,1.2083488);
\node[font=\tiny\sffamily, align=center, inner sep=1pt] at (1.5977260,1.2083488) {Osvaldo Mendoza};
\draw[dashed, gray!50, line width=0.3pt] (0.7500000,4.2500000) -- (1.2499569,2.6119088);
\node[font=\tiny\sffamily, align=center, inner sep=1pt] at (1.2499569,2.6119088) {Lee Dong-won};
\draw[dashed, gray!50, line width=0.3pt] (1.8194728,4.2500000) -- (2.1538768,2.2840603);
\node[font=\tiny\sffamily, align=center, inner sep=1pt] at (2.1538768,2.2840603) {Saša Simonović};
\draw[dashed, gray!50, line width=0.3pt] (1.8194728,4.2500000) -- (1.2945412,3.2633315);
\node[font=\tiny\sffamily, align=center, inner sep=1pt] at (1.2945412,3.2633315) {Jimmy Mulisa};
\draw[dashed, gray!50, line width=0.3pt] (1.8194728,4.2500000) -- (2.1899510,3.5941350);
\node[font=\tiny\sffamily, align=center, inner sep=1pt] at (2.1899510,3.5941350) {Karim Bagheri};
\draw[dashed, gray!50, line width=0.3pt] (1.8211982,4.2500000) -- (6.4459645,1.2781522);
\node[font=\tiny\sffamily, align=center, inner sep=1pt] at (6.4459645,1.2781522) {Macedonia national under-17 football team};
\draw[dashed, gray!50, line width=0.3pt] (1.8194728,4.2500000) -- (5.7019451,8.0000000);
\node[font=\tiny\sffamily, align=center, inner sep=1pt] at (5.7019451,8.0000000) {New Zealand women's national under-20 football team};
\draw[dashed, gray!50, line width=0.3pt] (1.8194728,4.2500000) -- (4.7416870,4.3963191);
\node[font=\tiny\sffamily, align=center, inner sep=1pt] at (4.7416870,4.3963191) {de/Deutsche Fußballnationalmannschaft (U-19-Frauen)};
\draw[dashed, gray!50, line width=0.3pt] (1.8194728,4.2500000) -- (4.3228582,0.0000000);
\node[font=\tiny\sffamily, align=center, inner sep=1pt] at (4.3228582,0.0000000) {United States women's national under-23 soccer team};
\draw[dashed, gray!50, line width=0.3pt] (7.7373680,4.2500000) -- (8.0000000,4.7768165);
\node[font=\tiny\sffamily, align=center, inner sep=1pt] at (8.0000000,4.7768165) {Thomas Bernhard};
\draw[dashed, gray!50, line width=0.3pt] (6.1915157,4.2500000) -- (7.5638596,2.8800730);
\node[font=\tiny\sffamily, align=center, inner sep=1pt] at (7.5638596,2.8800730) {Martin Buber};
\draw[dashed, gray!50, line width=0.3pt] (5.9747529,4.2500000) -- (6.6171434,5.0763017);
\node[font=\tiny\sffamily, align=center, inner sep=1pt] at (6.6171434,5.0763017) {Peter Kropotkin};
\draw[dashed, gray!50, line width=0.3pt] (7.7373680,4.2500000) -- (8.0000000,6.2050922);
\node[font=\tiny\sffamily, align=center, inner sep=1pt] at (8.0000000,6.2050922) {Battle of Marignano};
\draw[dashed, gray!50, line width=0.3pt] (7.7499953,7.7500000) -- (8.0000000,8.0000000);
\node[font=\tiny\sffamily, align=center, inner sep=1pt] at (8.0000000,8.0000000) {Josip Broz Tito};
\draw[dashed, gray!50, line width=0.3pt] (6.1915157,4.2500000) -- (7.1737242,4.2381227);
\node[font=\tiny\sffamily, align=center, inner sep=1pt] at (7.1737242,4.2381227) {Odessa};
\draw[dashed, gray!50, line width=0.3pt] (5.9747529,4.2500000) -- (6.4928916,3.4473695);
\node[font=\tiny\sffamily, align=center, inner sep=1pt] at (6.4928916,3.4473695) {Greater St. Louis};
\draw[dashed, gray!50, line width=0.3pt] (7.7373680,4.2500000) -- (8.0000000,3.5975499);
\node[font=\tiny\sffamily, align=center, inner sep=1pt] at (8.0000000,3.5975499) {Isaac Newton};
\draw[dashed, gray!50, line width=0.3pt] (6.1915157,4.2500000) -- (6.8734571,6.2222285);
\node[font=\tiny\sffamily, align=center, inner sep=1pt] at (6.8734571,6.2222285) {Ramon Magsaysay Award};
\draw[dashed, gray!50, line width=0.3pt] (7.7500000,0.7500000) -- (8.0000000,0.8859834);
\node[font=\tiny\sffamily, align=center, inner sep=1pt] at (8.0000000,0.8859834) {Rajinikanth};
\end{axis}
\end{tikzpicture}
}

\caption{Visualization of last layer embeddings (using PCA) for the ontology-enhanced model of \citet{zhou-etal-2023-inductive} for 50 candidate entities when predicting the missing tail for triple <\texttt{Andrei Gashkin, playsFor, ?}>. For all the 50 candidates, there is no enclosing subgraph.}

\label{fig:emb_visualization_zhou}

\end{figure*}

\textbf{\cref{fig:emb_visualization_zhou}} visualizes the PCA-projected embeddings from the model by~\citet{zhou-etal-2023-inductive}. In this visualization:
\begin{itemize}
    \item The correct tail entity, \texttt{FC KAMAZ Naberezhnye Chelny} (highlighted or labeled distinctly if possible in the actual figure), is positioned among a cluster of other football clubs and sports-related entities. For instance, it might be spatially close to other entities like \texttt{SV Grödig} or \texttt{Egri FC} if they were among the candidates.
    \item The embeddings of many semantically similar entities (e.g., various football clubs) are tightly clustered. This suggests that while the explicit type information used by this model (e.g., "Football Club" type) helps group entities by their broad category, it may not provide sufficient fine-grained discriminative power in this structurally sparse scenario.
    \item The model appears to struggle to clearly distinguish \texttt{FC KAMAZ Naberezhnye Chelny} from other plausible (same-type) but incorrect candidate entities based solely on the explicit type signals and the limited structural information available in the sparse subgraph. The representation reflects a general categorical understanding rather than a nuanced, context-specific one for the \texttt{playsFor} relation with \texttt{Andrei Gashkin}.
\end{itemize}

\begin{figure*}
\rotatebox{90}{%
\centering
\begin{tikzpicture}
\begin{axis}[
    width=1.4\textwidth,
    height=0.65\textheight,
    xlabel={$x$},
    ylabel={$y$},
    xmin=-1, xmax=9,
    ymin=-1, ymax=9,
    grid=none,
    grid style={line width=.1pt, draw=gray!30},
    major grid style={line width=.2pt, draw=gray!50},
    minor tick num=4,
    tick style={black, line width=0.5pt},
    axis line style={-latex},
    xlabel style={at={(axis description cs:1.02,0.02)}, anchor=west, font=\small\sffamily},
    ylabel style={rotate=-90, at={(axis description cs:0.02,1.02)}, anchor=south, font=\small\sffamily},
    ticklabel style={font=\small\sffamily},
    clip=false,
    enlarge x limits = false
]
\addplot[mark=*, mark size=1.5pt, color=red] coordinates {(7.2057222,2.1948513)};
\addplot[mark=*, mark size=1.5pt, color=red] coordinates {(0.7551022,2.9443126)};
\addplot[mark=*, mark size=1.5pt, color=red] coordinates {(7.7500000,0.7500000)};
\addplot[mark=*, mark size=1.5pt, color=red] coordinates {(0.7509980,2.9462088)};
\addplot[mark=*, mark size=1.5pt, color=red] coordinates {(7.2068278,2.4598486)};
\addplot[mark=*, mark size=1.5pt, color=red] coordinates {(7.2090473,2.4508262)};
\addplot[mark=*, mark size=1.5pt, color=red] coordinates {(0.7505336,2.9459533)};
\addplot[mark=*, mark size=1.5pt, color=red] coordinates {(5.0324927,4.1242730)};
\addplot[mark=*, mark size=1.5pt, color=red] coordinates {(7.2091482,2.4510146)};
\addplot[mark=*, mark size=1.5pt, color=red] coordinates {(0.7510938,2.9455412)};
\addplot[mark=*, mark size=1.5pt, color=red] coordinates {(3.8406047,3.5837169)};
\addplot[mark=*, mark size=1.5pt, color=red] coordinates {(7.7195278,0.8563582)};
\addplot[mark=*, mark size=1.5pt, color=red] coordinates {(0.7542978,2.9481625)};
\addplot[mark=*, mark size=1.5pt, color=red] coordinates {(0.7522212,2.9449726)};
\addplot[mark=*, mark size=1.5pt, color=red] coordinates {(0.7500000,2.9459433)};
\addplot[mark=*, mark size=1.5pt, color=red] coordinates {(0.7554375,2.9480678)};
\addplot[mark=*, mark size=1.5pt, color=red] coordinates {(4.4279982,3.3281534)};
\addplot[mark=*, mark size=1.5pt, color=red] coordinates {(0.7547811,2.9485862)};
\addplot[mark=*, mark size=1.5pt, color=red] coordinates {(0.7501926,2.9455351)};
\addplot[mark=*, mark size=1.5pt, color=red] coordinates {(7.2096333,2.4489793)};
\addplot[mark=*, mark size=1.5pt, color=red] coordinates {(0.7550818,2.9483889)};
\addplot[mark=*, mark size=1.5pt, color=red] coordinates {(7.2091651,2.4504387)};
\addplot[mark=*, mark size=1.5pt, color=red] coordinates {(0.7503944,2.9459524)};
\addplot[mark=*, mark size=1.5pt, color=red] coordinates {(4.9590696,7.7500000)};
\addplot[mark=*, mark size=1.5pt, color=red] coordinates {(7.2087755,2.4512690)};
\addplot[mark=*, mark size=1.5pt, color=red] coordinates {(4.9650586,7.2856782)};
\addplot[mark=*, mark size=1.5pt, color=red] coordinates {(0.7562352,2.9491792)};
\addplot[mark=*, mark size=1.5pt, color=red] coordinates {(5.7217156,7.2112981)};
\addplot[mark=*, mark size=1.5pt, color=red] coordinates {(0.7517911,2.9452524)};
\addplot[mark=*, mark size=1.5pt, color=red] coordinates {(0.7513674,2.9452510)};
\addplot[mark=*, mark size=1.5pt, color=red] coordinates {(0.7511185,2.9462512)};
\addplot[mark=*, mark size=1.5pt, color=red] coordinates {(0.7500105,2.9455250)};
\addplot[mark=*, mark size=1.5pt, color=red] coordinates {(0.7503012,2.9459523)};
\addplot[mark=*, mark size=1.5pt, color=red] coordinates {(0.7556105,2.9482088)};
\addplot[mark=*, mark size=1.5pt, color=red] coordinates {(0.7534059,2.9478327)};
\addplot[mark=*, mark size=1.5pt, color=red] coordinates {(7.2090568,2.4507653)};
\addplot[mark=*, mark size=1.5pt, color=red] coordinates {(0.7523365,2.9464896)};
\addplot[mark=*, mark size=1.5pt, color=red] coordinates {(0.7515530,2.9460082)};
\addplot[mark=*, mark size=1.5pt, color=red] coordinates {(1.0477203,2.9239913)};
\addplot[mark=*, mark size=1.5pt, color=red] coordinates {(0.7566811,2.9496224)};
\addplot[mark=*, mark size=1.5pt, color=red] coordinates {(3.8411762,3.5895626)};
\addplot[mark=*, mark size=1.5pt, color=red] coordinates {(0.7500578,2.9457895)};
\addplot[mark=*, mark size=1.5pt, color=red] coordinates {(0.7506789,2.9457329)};
\addplot[mark=*, mark size=1.5pt, color=red] coordinates {(0.7586063,2.9434793)};
\addplot[mark=*, mark size=1.5pt, color=red] coordinates {(4.8777215,7.0136280)};
\addplot[mark=*, mark size=1.5pt, color=red] coordinates {(7.1756013,2.5879747)};
\addplot[mark=*, mark size=1.5pt, color=red] coordinates {(0.7539127,2.9449267)};
\addplot[mark=*, mark size=1.5pt, color=red] coordinates {(7.2113299,2.4437542)};
\addplot[mark=*, mark size=1.5pt, color=red] coordinates {(0.7692138,2.9565112)};
\addplot[mark=*, mark size=1.5pt, color=red] coordinates {(0.7519304,2.9454630)};
\draw[dashed, gray!50, line width=0.3pt] (7.2113299,2.1837542) -- (5.1395652,0.0000000);
\node[font=\tiny\sffamily, align=center, inner sep=1pt, text=green!60!black] at (5.1395652,0.0000000) {FC KAMAZ Naberezhnye Chelny};
\draw[dashed, gray!50, line width=0.3pt] (1.0477203,2.9239913) -- (3.0586203,0.0423333);
\node[font=\tiny\sffamily, align=center, inner sep=1pt] at (3.0586203,0.0423333) {Rafał Berliński};
\draw[dashed, gray!50, line width=0.3pt] (7.7195278,0.8563582) -- (8.0000000,0.0000000);
\node[font=\tiny\sffamily, align=center, inner sep=1pt] at (8.0000000,0.0000000) {Atlético Petróleos de Luanda};
\draw[dashed, gray!50, line width=0.3pt] (1.0477203,2.9239913) -- (0.8502352,5.4589425);
\node[font=\tiny\sffamily, align=center, inner sep=1pt] at (0.8502352,5.4589425) {Boris Rotenberg};
\draw[dashed, gray!50, line width=0.3pt] (7.1756013,2.5879747) -- (6.0931370,4.0000000);
\node[font=\tiny\sffamily, align=center, inner sep=1pt] at (6.0931370,4.1000000) {Juventud Independiente};
\draw[dashed, gray!50, line width=0.3pt] (7.1756013,2.5879747) -- (4.8993340,1.4787233);
\node[font=\tiny\sffamily, align=center, inner sep=1pt] at (4.8993340,1.4087233) {Macedonia national under-17 football team};
\draw[dashed, gray!50, line width=0.3pt] (1.0477203,2.9239913) -- (0.7914472,3.8617215);
\node[font=\tiny\sffamily, align=center, inner sep=1pt] at (0.7914472,3.8617215) {Mbo Mpenza};
\draw[dashed, gray!50, line width=0.3pt] (5.0324927,4.1242730) -- (5.2942916,5.0988016);
\node[font=\tiny\sffamily, align=center, inner sep=1pt] at (5.2942916,5.0988016) {León de Huánuco};
\draw[dashed, gray!50, line width=0.3pt] (7.1756013,2.5879747) -- (7.4000000,7.8000000);
\node[font=\tiny\sffamily, align=center, inner sep=1pt] at (7.4000000,7.8000000) {New Zealand women's national under-20 football team};
\draw[dashed, gray!50, line width=0.3pt] (1.0477203,2.9239913) -- (1.1536809,1.6826214);
\node[font=\tiny\sffamily, align=center, inner sep=1pt] at (1.1536809,1.6826214) {Uladzimir Haew};
\draw[dashed, gray!50, line width=0.3pt] (3.8411762,3.5895626) -- (4.3546445,3.1077460);
\node[font=\tiny\sffamily, align=center, inner sep=1pt] at (4.3546445,3.1077460) {Rimas Kurtinaitis};
\draw[dashed, gray!50, line width=0.3pt] (7.7195278,0.8563582) -- (6.4757972,0.0000000);
\node[font=\tiny\sffamily, align=center, inner sep=1pt] at (6.4757972,0.0000000) {Egri FC};
\draw[dashed, gray!50, line width=0.3pt] (1.0477203,2.9239913) -- (1.5121014,4.6928299);
\node[font=\tiny\sffamily, align=center, inner sep=1pt] at (1.5121014,4.6928299) {Thomas Bernhard};
\draw[dashed, gray!50, line width=0.3pt] (1.0477203,2.9239913) -- (0.0000000,0.1500000);
\node[font=\tiny\sffamily, align=center, inner sep=1pt] at (0.2500000,0.3500000) {Jacek Zieliński (footballer born 1961)};
\draw[dashed, gray!50, line width=0.3pt] (1.0477203,2.9239913) -- (0.0000000,5.1136180);
\node[font=\tiny\sffamily, align=center, inner sep=1pt] at (0.0000000,5.1136180) {Herfried Sabitzer};
\draw[dashed, gray!50, line width=0.3pt] (1.0477203,2.9239913) -- (3.9943315,4.4386502);
\node[font=\tiny\sffamily, align=center, inner sep=1pt] at (3.9943315,4.4386502) {Jamie Arnold (basketball)};
\draw[dashed, gray!50, line width=0.3pt] (4.4279982,3.3281534) -- (5.1151823,3.7959955);
\node[font=\tiny\sffamily, align=center, inner sep=1pt] at (5.1151823,3.7959955) {FK Teteks};
\draw[dashed, gray!50, line width=0.3pt] (1.0477203,2.9239913) -- (1.5042936,6.0963198);
\node[font=\tiny\sffamily, align=center, inner sep=1pt] at (1.5042936,6.0963198) {Byron Dinkins};
\draw[dashed, gray!50, line width=0.3pt] (1.0477203,2.9239913) -- (0.0000000,1.3700258);
\node[font=\tiny\sffamily, align=center, inner sep=1pt] at (0.0000000,1.3700258) {Arman Karamyan};
\draw[dashed, gray!50, line width=0.3pt] (7.1756013,2.5879747) -- (8.0000000,3.1420051);
\node[font=\tiny\sffamily, align=center, inner sep=1pt] at (8.0000000,3.1420051) {Stationery Stores F.C.};
\draw[dashed, gray!50, line width=0.3pt] (1.0477203,2.9239913) -- (2.4758631,3.8252990);
\node[font=\tiny\sffamily, align=center, inner sep=1pt] at (2.4758631,3.8252990) {Martin Buber};
\draw[dashed, gray!50, line width=0.3pt] (7.1756013,2.5879747) -- (7.5000000,5.4295444);
\node[font=\tiny\sffamily, align=center, inner sep=1pt] at (7.5000000,5.4295444) {de/Deutsche Fußballnationalmannschaft (U-19-Frauen)};
\draw[dashed, gray!50, line width=0.3pt] (1.0477203,2.9239913) -- (0.0000000,3.5583337);
\node[font=\tiny\sffamily, align=center, inner sep=1pt] at (0.0000000,3.5583337) {Marcel Schied};
\draw[dashed, gray!50, line width=0.3pt] (4.9590696,7.7500000) -- (4.2981836,8.0000000);
\node[font=\tiny\sffamily, align=center, inner sep=1pt] at (4.2981836,8.0000000) {Battle of Marignano};
\draw[dashed, gray!50, line width=0.3pt] (7.1756013,2.5879747) -- (6.1194022,2.4487321);
\node[font=\tiny\sffamily, align=center, inner sep=1pt] at (6.1194022,2.4487321) {Dohuk SC};
\draw[dashed, gray!50, line width=0.3pt] (4.9590696,7.3000000) -- (5.7056287,6.8587525);
\node[font=\tiny\sffamily, align=center, inner sep=1pt] at (5.7056287,6.8587525) {Valencia BC};
\draw[dashed, gray!50, line width=0.3pt] (1.0477203,2.9239913) -- (2.4088176,5.4777780);
\node[font=\tiny\sffamily, align=center, inner sep=1pt] at (2.4088176,5.4777780) {Josip Broz Tito};
\draw[dashed, gray!50, line width=0.3pt] (5.7217156,7.2112981) -- (5.7125998,8.0000000);
\node[font=\tiny\sffamily, align=center, inner sep=1pt] at (5.7125998,8.0000000) {Odessa};
\draw[dashed, gray!50, line width=0.3pt] (1.0477203,2.9239913) -- (1.7138074,0.0000000);
\node[font=\tiny\sffamily, align=center, inner sep=1pt] at (1.7138074,0.0000000) {Zoltán Balog (footballer)};
\draw[dashed, gray!50, line width=0.3pt] (1.0477203,2.9239913) -- (1.9514361,1.2156416);
\node[font=\tiny\sffamily, align=center, inner sep=1pt] at (1.9514361,1.2156416) {Roman Uzdenov};
\draw[dashed, gray!50, line width=0.3pt] (1.0477203,2.9239913) -- (0.0000000,6.5386668);
\node[font=\tiny\sffamily, align=center, inner sep=1pt] at (0.0000000,6.5386668) {José Emilio Furtado};
\draw[dashed, gray!50, line width=0.3pt] (1.0477203,2.9239913) -- (0.0000000,2.1147986);
\node[font=\tiny\sffamily, align=center, inner sep=1pt] at (0.0000000,2.1147986) {Róbert Zsolnai};
\draw[dashed, gray!50, line width=0.3pt] (1.0477203,2.9239913) -- (0.0000000,4.2425563);
\node[font=\tiny\sffamily, align=center, inner sep=1pt] at (0.0000000,4.2425563) {Ivan Hodúr};
\draw[dashed, gray!50, line width=0.3pt] (1.0477203,2.9239913) -- (2.9645456,3.2887533);
\node[font=\tiny\sffamily, align=center, inner sep=1pt] at (2.9645456,3.2887533) {Jamal Mohammed};
\draw[dashed, gray!50, line width=0.3pt] (1.0477203,2.9239913) -- (1.2548843,7.6550947);
\node[font=\tiny\sffamily, align=center, inner sep=1pt] at (1.2548843,7.6550947) {Luciano Valente de Deus};
\draw[dashed, gray!50, line width=0.3pt] (7.1756013,2.5879747) -- (6.6585284,2.9747764);
\node[font=\tiny\sffamily, align=center, inner sep=1pt] at (6.6585284,2.9747764) {SV Grödig};
\draw[dashed, gray!50, line width=0.3pt] (1.0477203,2.9239913) -- (1.6209203,4.0909441);
\node[font=\tiny\sffamily, align=center, inner sep=1pt] at (1.6209203,4.1099441) {André Martins};
\draw[dashed, gray!50, line width=0.3pt] (1.0477203,2.9239913) -- (1.8232794,3.0287226);
\node[font=\tiny\sffamily, align=center, inner sep=1pt] at (1.8232794,3.0287226) {Muharem Bajrami};
\draw[dashed, gray!50, line width=0.3pt] (1.0477203,2.9239913) -- (3.2633973,2.6086055);
\node[font=\tiny\sffamily, align=center, inner sep=1pt] at (3.2633973,2.6086055) {Isaac Newton};
\draw[dashed, gray!50, line width=0.3pt] (1.0477203,2.9239913) -- (2.7631754,6.8163382);
\node[font=\tiny\sffamily, align=center, inner sep=1pt] at (2.7631754,6.8163382) {Peter Kropotkin};
\draw[dashed, gray!50, line width=0.3pt] (3.8411762,3.5895626) -- (3.8466501,5.8793033);
\node[font=\tiny\sffamily, align=center, inner sep=1pt] at (3.8466501,5.8793033) {Ramon Magsaysay Award};
\draw[dashed, gray!50, line width=0.3pt] (1.0477203,2.9239913) -- (0.0000000,2.8486884);
\node[font=\tiny\sffamily, align=center, inner sep=1pt] at (0.0000000,2.8486884) {Osvaldo Mendoza};
\draw[dashed, gray!50, line width=0.3pt] (1.0477203,2.9239913) -- (0.8757248,2.5588834);
\node[font=\tiny\sffamily, align=center, inner sep=1pt] at (0.8757248,2.5588834) {Lee Dong-won};
\draw[dashed, gray!50, line width=0.3pt] (1.0477203,2.9239913) -- (3.2057124,1.2684096);
\node[font=\tiny\sffamily, align=center, inner sep=1pt] at (3.2057124,1.2684096) {Saša Simonović};
\draw[dashed, gray!50, line width=0.3pt] (4.8550586,6.9556782) -- (4.7606468,6.6203826);
\node[font=\tiny\sffamily, align=center, inner sep=1pt] at (4.7606468,6.6203826) {Greater St. Louis};
\draw[dashed, gray!50, line width=0.3pt] (7.1756013,2.5879747) -- (5.9868455,5.9814999);
\node[font=\tiny\sffamily, align=center, inner sep=1pt] at (5.9868455,5.9814999) {Dakota Wizards};
\draw[dashed, gray!50, line width=0.3pt] (1.0477203,2.9239913) -- (2.5698856,1.9084648);
\node[font=\tiny\sffamily, align=center, inner sep=1pt] at (2.5698856,1.9084648) {Jimmy Mulisa};
\draw[dashed, gray!50, line width=0.3pt] (7.1756013,2.5879747) -- (7.6000000,1.3466724);
\node[font=\tiny\sffamily, align=center, inner sep=1pt] at (7.6000000,1.3466724) {United States women's national under-23 soccer team};
\draw[dashed, gray!50, line width=0.3pt] (1.0477203,2.9239913) -- (2.7699291,4.6500906);
\node[font=\tiny\sffamily, align=center, inner sep=1pt] at (2.7699291,4.6500906) {Rajinikanth};
\draw[dashed, gray!50, line width=0.3pt] (1.0477203,2.9239913) -- (2.0589115,2.4916939);
\node[font=\tiny\sffamily, align=center, inner sep=1pt] at (2.0589115,2.4916939) {Karim Bagheri};
\end{axis}
\end{tikzpicture}
}

\caption{Visualization of last layer embeddings (using PCA) for \tyler (\robertal) for 50 candidate entities when predicting the missing tail for triple <\texttt{Andrei Gashkin, playsFor, ?}>. For all the 50 candidates, there is no enclosing subgraph.}

\label{fig:emb_visualization_tyler}

\end{figure*}
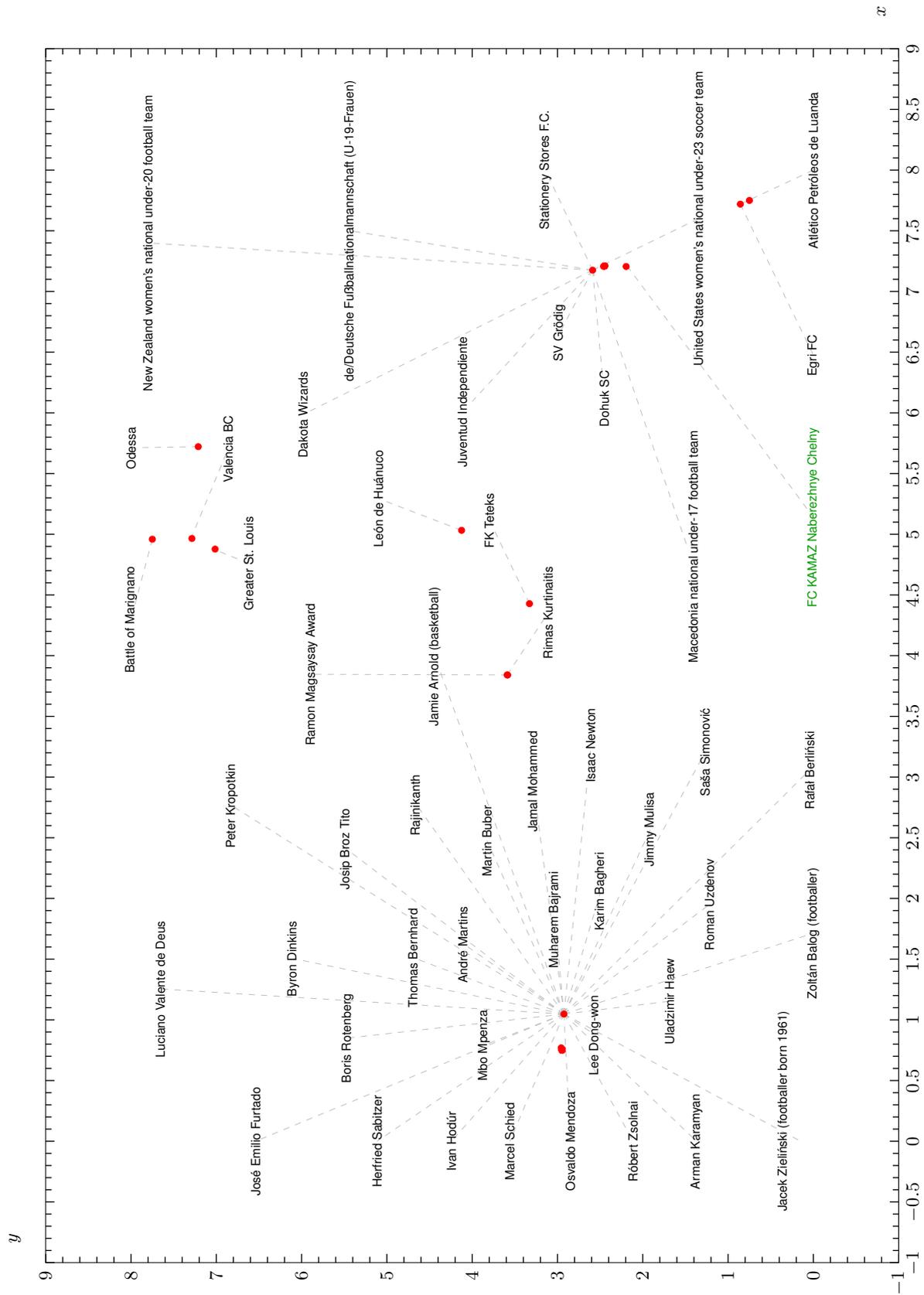

\noindent \textbf{\cref{fig:emb_visualization_tyler}} displays the PCA-projected embeddings from our \tyler-\robertal model for the same set of 50 candidate entities.
 \begin{itemize}
     \item The correct tail entity, \texttt{FC KAMAZ Naberezhnye Chelny}, is noticeably more separated in the embedding space compared to its representation in \cref{fig:emb_visualization_zhou}. While it would still likely be in a region associated with sports entities, its position relative to other incorrect candidate football clubs is more distinct.
     \item This improved separation suggests that \tyler's semantic enrichment, derived from \robertal, provides more nuanced and discriminative features. The model benefits from the implicit propagation of semantic information related to the head entity \texttt{Andrei Gashkin} (a known footballer) through the PLM's understanding.
     \item The PLM's pre-trained knowledge helps infer a more fine-grained "type-awareness" and contextual understanding for the \texttt{playsFor} relation. Even with sparse explicit graph structure, \tyler can leverage the rich semantics encoded by the PLM (and potentially GNN mechanisms like self-loop connections that reinforce entity identity) to better characterize and differentiate the correct tail entity.
 \end{itemize}

\textit{This visual comparison underscores the benefit of \tyler’s approach in handling structurally sparse scenarios. The ontology-enhanced model (\citet{zhou-etal-2023-inductive}), while utilizing explicit types, produces less distinguishable embeddings for semantically similar entities when graph structure is poor. In contrast, \tyler, by incorporating rich implicit type signals from a pre-trained language model, achieves a more fine-grained characterization and better separation of the correct entity in the embedding space. This highlights the potential of PLM-derived semantic enrichment to compensate for deficiencies in explicit type annotations and structural connectivity, leading to more robust inductive link prediction. This supports our paper's argument that implicit type signals enable a more nuanced understanding, particularly crucial in sparse settings.}

\end{document}